\def\FtwoGArxiv{1}
\documentclass{article}

\PassOptionsToPackage{numbers,sort&compress}{natbib}
\ifdefined\FtwoGArxiv
  \usepackage[preprint]{neurips_2026}
\else
  \usepackage[main,final]{neurips_2026}
\fi

\usepackage[utf8]{inputenc} 
\usepackage[T1]{fontenc}    

\usepackage{url}            
\usepackage{booktabs}       
\usepackage{amsfonts}       
\usepackage{nicefrac}       
\usepackage{microtype}      
\usepackage{xcolor}         
\usepackage{graphicx}
\usepackage{wrapfig}
\usepackage{amsmath}
\usepackage{multirow}
\usepackage{caption}

\definecolor{urlblue}{RGB}{40,120,210}

\usepackage{hyperref}
\hypersetup{
    colorlinks=true,
    urlcolor=urlblue,
    linkcolor=urlblue,
    citecolor=urlblue
}

\makeatletter
\renewcommand\section{\@startsection{section}{1}{\z@}%
  {-1.35ex plus -.3ex minus -.2ex}%
  {0.55ex plus .15ex}%
  {\normalfont\Large\bfseries}}
\renewcommand\subsection{\@startsection{subsection}{2}{\z@}%
  {-0.9ex plus -.25ex minus -.15ex}%
  {0.3ex plus .15ex}%
  {\normalfont\large\bfseries}}
\makeatother

\usepackage{xcolor}

\newcommand{\rev}[1]{\textcolor{black}{#1}} 

\title{F2G-Pose: Geometry-Aware \\ Foundation Feature Lifting for Direct RGB-D \\ Category-Level Object Pose Estimation}

\author{%
  Jinyu Zhang\textsuperscript{1,2}\thanks{Equal contribution. Contact: \href{mailto:zhangjinyu25@sii.edu.cn}{zhangjinyu25@sii.edu.cn}.}\quad
  Haitao Lin\textsuperscript{1}\footnotemark[1]\quad
  Jiashu Hou\textsuperscript{1}\footnotemark[1]\\
  \textbf{Xiangyang Xue}\textsuperscript{1}\quad
  \textbf{Yanwei Fu}\textsuperscript{1,2}\thanks{Corresponding author. Contact: \href{mailto:yanweifu@fudan.edu.cn}{yanweifu@fudan.edu.cn}.}\\
  \textsuperscript{1}Fudan University\qquad
  \textsuperscript{2}Shanghai Innovation Institute
}

\begin{document}

\maketitle

\begin{abstract}
\rev{This paper studies category-level RGB-D object pose estimation, which recovers an object's 3D rotation and translation without instance-specific CAD models or reference views. This task is highly ambiguous due to intra-class variations, symmetries, and partial observations. Current solutions struggle to balance reliability and efficiency: methods based on intermediate structures are prone to error propagation, while diffusion models suffer from high inference latency and complex post-hoc candidate extraction. }
To address this, we present F2G-Pose, a real-time direct, single-pass framework that predicts category-level pose, metric size, and camera-frame completed shape from a single RGB-D observation in one feed-forward pass. F2G-Pose lifts dense visual foundation features to partial point clouds, encodes local geometry with DGCNN tokenization, and integrates object-level structure with a Mixture-of-Experts Transformer. Joint size prediction and camera-frame shape completion encourage metric consistency through dense geometric supervision. With its task-specific network trained only on pose-annotated synthetic SOPE data and its frozen RADIO backbone externally pretrained, F2G-Pose achieves state-of-the-art accuracy on SOPE and ROPE while running at 28 FPS for pose-only inference, and further demonstrates strong camera-frame geometry on HouseCat6D and real-object transfer on HANDAL. Analyses of latency, robustness, shape completion, and GenPose++ candidate behavior further demonstrate the efficiency and reliability of F2G-Pose, with real-world manipulation experiments demonstrating deployment potential under segmented RGB-D inputs.
The project page, code, and pretrained models are made available at \url{https://jrryzh.github.io/F2G-Pose/}.
\end{abstract}


\section{Introduction}

Category-level 6D object pose estimation recovers an object's 3D rotation and translation without requiring an instance-specific CAD model or reference views at test time. In practical RGB-D perception, this estimate is often coupled with metric size and usable object geometry for benchmark evaluation~\cite{wang2019normalized,zhang2025omni6dpose,fu2022category}, physical interaction~\cite{tan20176d,tang20193d,durchon20246d}, and robotic control~\cite{wen2023bundlesdf,lin2023pourit,wang2024polaris,huang2025cap,huang2025you,lin2026universal}.
\rev{This setting is inherently challenging because partial observations from a single crop or segmented point cloud provide incomplete evidence: symmetries induce rotational ambiguity, occlusions hide discriminative geometry, and intra-class variations alter shape, size, and appearance.}
\rev{Consequently, category-level perception must produce reliable metric predictions from ambiguous, incomplete, and noisy observations.}

Previous work has addressed this problem from two complementary directions. Deterministic category-level methods estimate pose and size through canonical coordinates~\cite{wang2019normalized}, correspondences~\cite{chen2021fs}, shape priors~\cite{irshad2022shapo}, deformation models~\cite{tian2020shape}, alignment procedures~\cite{lin2022sar}, or direct regression~\cite{liu2023net,wang2022phocal,jung2024housecat6d}. 
\rev{While these methods enable efficient inference, reliance on intermediate structures can lead to compounding errors.}
Under occlusion, missing depth, intra-class variation, and domain shift, the final metric estimate can become sensitive to the reliability of these intermediate structures. Direct regression can avoid some post-hoc alignment steps, but reliable direct prediction still requires representations that preserve local geometric cues and global object structure under partial observations.

A complementary line of work models pose ambiguity explicitly. GenPose~\cite{zhang2023genpose} and GenPose++~\cite{zhang2025omni6dpose} formulate category-level pose estimation as conditional pose distribution modeling, sampling multiple hypotheses and converting them into a prediction through scoring, filtering, clustering, or aggregation. This is principled for uncertainty representation, but it introduces a candidate-to-decision gap: deployed accuracy depends not only on the sampled hypotheses, but also on whether post-hoc extraction can select or aggregate a reliable representative. Moreover, this conversion must operate under a practical inference budget as in Fig.~\ref{fig:pareto_front}, so sampling quality, candidate extraction, and computational cost become coupled. Joint-RegDiff~\cite{lee2025jointlearningposeregression} and RFMPose~\cite{ouyang2025rfmpose} improve generative pose inference through joint regression--diffusion learning with score-scaling sampling and Riemannian flow-matching trajectories, respectively, but both still rely on iterative generative inference to produce the prediction, so prediction quality remains coupled with
sampling dynamics, candidate extraction, and inference cost.

\rev{We introduce F2G-Pose, a direct RGB-D framework that jointly estimates category-level pose, size, and camera-frame completed geometry in a single feed-forward pass. By leveraging geometry-aware foundation feature lifting and dense geometric supervision, it bypasses intermediate canonical shape alignment and avoids the sampling and candidate extraction used by sampling-based pose estimators.}
\rev{Specifically, we lift dense visual foundation features~\cite{heinrich2025radiov2} to the partial point cloud for semantic cues, encode local geometry via DGCNN tokenization, and integrate object-level structure using a Mixture-of-Experts Transformer that provides conditional capacity across diverse shape regimes.}
Then, the pose-size head predicts the final metric estimate, while the shape head reconstructs camera-frame geometry as both auxiliary supervision and a usable output. This joint design encourages the shared representation to encode visible local evidence, object extent, and visible-to-complete geometric structure for consistent direct prediction under partial observations.

To evaluate the effectiveness of this design, we train the task-specific network only on pose-annotated synthetic data from 149 SOPE categories~\cite{zhang2025omni6dpose} and test on SOPE~\cite{zhang2025omni6dpose}, ROPE~\cite{zhang2025omni6dpose}, HANDAL~\cite{guo2023handal}, and HouseCat6D~\cite{jung2024housecat6d}. These benchmarks cover synthetic benchmark accuracy, synthetic-to-real transfer, additional real-object evaluation, and camera-frame shape reconstruction. F2G-Pose achieves state-of-the-art accuracy while running at 28 FPS for pose inference. Beyond benchmark accuracy, we evaluate latency, robustness to perturbations and occlusion, shape reconstruction, GenPose++ candidate behavior, and real-world robotic manipulation.
\rev{These results demonstrate the efficiency and reliability of direct feed-forward prediction, while robotic experiments demonstrate deployment potential under segmented RGB-D inputs.}

\noindent \textbf{Contributions.} Our contributions are threefold:
\textbf{(1)} 
We present F2G-Pose, a real-time direct RGB-D framework that predicts category-level 6D pose, metric size, and camera-frame completed shape in a single feed-forward pass, sharing its representation across pose, size, and shape learning. 
\textbf{(2)} 
\rev{We introduce geometry-aware foundation feature lifting to map dense visual features onto partial point clouds. DGCNN encodes local geometry, a Mixture-of-Experts Transformer integrates object-level structure, and camera-frame shape completion provides dense geometric supervision for metric prediction.}
\textbf{(3)}
\rev{Extensive evaluations across synthetic benchmarks and real-world datasets demonstrate that F2G-Pose achieves state-of-the-art accuracy, real-time inference, and high-quality shape completion. Furthermore, physical manipulation experiments demonstrate pick-and-place success, timely decision readiness, and deployment feasibility under segmented RGB-D inputs.}

\section{Related work}

\noindent\textbf{Category-Level 6D Pose Estimation.}
Category-level methods aim to generalize pose estimation across unseen object instances within a category~\cite{wang2019normalized,wang2022phocal,jung2024housecat6d}. Early works such as NOCS~\cite{wang2019normalized} introduced the notion of canonical space to enable pose alignment without requiring CAD models. Follow-up approaches~\cite{lin2022sar,liu2023net,chen2021fs,zhang2023genpose} leveraged point cloud geometry and symmetry-aware losses to improve generalization. Some methods further incorporate shape-aware modeling~\cite{tian2020shape,irshad2022shapo} to jointly predict object size and shape.
While these methods remove the need for object-specific models, many rely on intermediate canonical coordinates, correspondences, deformation fields, or shape priors whose reliability can degrade under occlusion, missing depth, and strong intra-category variation. Recent works have attempted to scale category-level 6D learning using larger and more diverse datasets~\cite{zhang2025omni6d,krishnan2024omninocs}, making representation quality and inference efficiency increasingly important.
Diffusion-based methods such as GenPose~\cite{zhang2023genpose} and GenPose++~\cite{zhang2025omni6dpose} model pose and size as multi-modal distributions, but often rely on auxiliary scoring networks to filter outlier samples.
Joint-RegDiff~\cite{lee2025jointlearningposeregression} eliminates the need for such scoring networks through time-dependent score scaling, and RFMPose~\cite{ouyang2025rfmpose} formulates generative category-level pose estimation with Riemannian flow matching; however, both remain limited to 6D pose estimation and still require iterative generative inference, which generally increases inference cost and can limit real-time throughput.

SecondPose~\cite{chen2024secondpose} also combines foundation-model features with geometry for direct category-level pose and size estimation through an SE(3)-consistent representation. F2G-Pose additionally predicts complete camera-frame shape to provide dense supervision for the shared pose-size representation, and uses MoE encoding to provide conditional capacity across diverse categories.

\noindent\textbf{Instance-Level Novel Object Pose Estimation.}
Another line of work estimates pose for novel object instances using CAD models~\cite{labbe2022megapose,nguyen2024gigapose}, single-view references, or multi-view references~\cite{lee2025any6d,liu2025one2any,he2022onepose++}. They often reconstruct object geometry, either explicitly via 3D modeling~\cite{liu2022gen6d,li2023nerf} or implicitly through image-based retrieval~\cite{he2022fs6d,nguyen2022templates}. FoundationPose~\cite{wen2024foundationpose} supports either CAD or reference images, combining model-based and model-free paradigms.
Recent foundation reconstruction models such as SAM 3D~\cite{sam3d2025} reconstruct object geometry, texture, and layout from masked single images, targeting image-to-3D asset reconstruction in natural scenes.
Together, these methods broaden object-centric 3D perception, but generally target instance-specific reconstruction or reference-conditioned alignment rather than category-level pose-size estimation under a fixed taxonomy.
F2G-Pose targets this complementary category-level setting: it uses category-level canonical supervision during training, but requires only a segmented RGB-D observation at inference time for the evaluated taxonomy.

\section{Method}
\noindent \textbf{Problem Formulation and Overview}
Given an RGB image crop \( \mathbf{I} \in \mathbb{R}^{H \times W \times 3} \) and a segmented partial point cloud \( \mathbf{P} \in \mathbb{R}^{N \times 3} \), F2G-Pose predicts the object's 6D pose \( \mathbf{T}=(\mathbf{R}, \mathbf{t}) \in \mathrm{SE}(3) \), metric size \( \mathbf{s} \in \mathbb{R}^3 \), and completed camera-frame shape \( \mathbf{P}_\text{dense} \in \mathbb{R}^{N_d \times 3} \) in a single feed-forward pass, without requiring instance-specific CAD models, reference images, or explicit category labels.
Here, \( \mathbf{R} \in \mathrm{SO}(3) \) and \( \mathbf{t} \in \mathbb{R}^3 \) denote the object rotation and translation, respectively.

\begin{figure*}[t]
    \centering
    \includegraphics[width=0.85\textwidth]{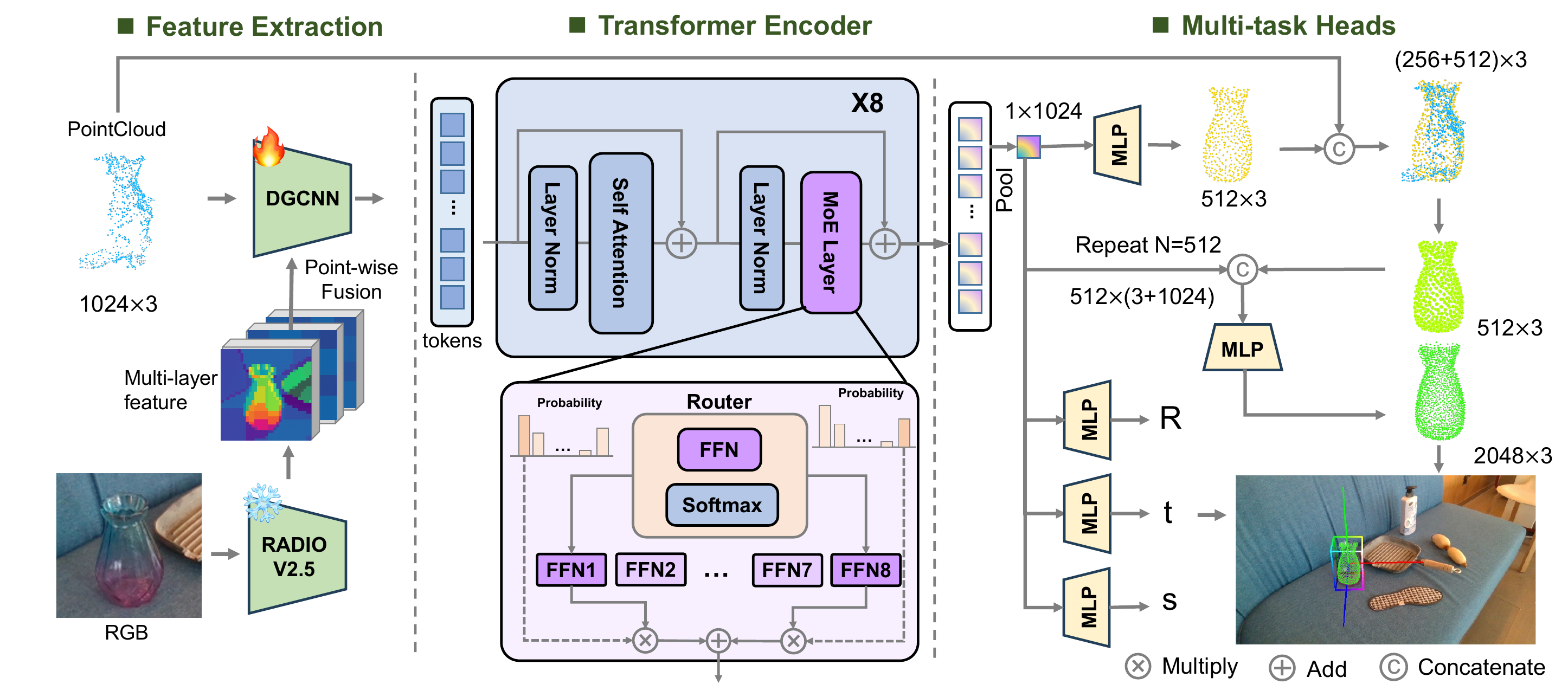}
    \caption{\textbf{F2G-Pose Overview.} Given a cropped RGB image and segmented partial point cloud, F2G-Pose extracts dense RADIOv2.5 features~\cite{heinrich2025radiov2} and lifts them onto 3D points by projection. DGCNN tokenization produces fused RGB-D object tokens, and a MoE Transformer encodes a global object representation. Two parallel decoder branches directly predict (i) 6D pose and metric size and (ii) camera-frame completed shape through coarse-to-fine refinement. The pipeline produces a pose-size-shape estimate in one feed-forward pass.
    }
    \label{fig:pipeline}
\end{figure*}

The framework directly predicts the outputs, avoiding the test-time sampling and candidate extraction of sampling-based estimators, as shown in Fig.~\ref{fig:pipeline}.
Instead, it lifts dense visual foundation features onto observed 3D points, encodes local geometric structure with DGCNN, aggregates object-level context with a MoE Transformer, and uses parallel decoder branches for pose-size prediction and camera-frame shape completion.
The pose-size branch directly outputs the metric object state used for evaluation and downstream applications, while the shape branch provides dense geometric supervision for the shared RGB-D representation and produces approximate completed geometry in the camera frame.

\subsection{Geometry-Aware Foundation Feature Lifting}

F2G-Pose uses RADIOv2.5~\cite{heinrich2025radiov2} to provide dense visual evidence for the partial RGB-D observation. RADIOv2.5 distills complementary 2D vision models~\cite{ravi2024sam,oquab2023dinov2,radford2021learning,fang2023data}, combining dense feature extraction from SAM~\cite{ravi2024sam} with semantic features from DINOv2~\cite{oquab2023dinov2}. Prior work has shown that DINOv2-style local features can support pose-consistent correspondences across object appearance and shape variation~\cite{zhang2024tale,zhang2024lapose}. We use these features not as a standalone image prior, but as point-wise appearance and semantic cues attached to metric 3D observations.
Given an RGB crop $\mathbf{I}$, we fuse RADIOv2.5 layers $l\in\{8,16,23\}$ as $\mathbf{F}_{\text{rgb}}=\mathcal{R}\big(\sum_l\alpha_l\mathcal{C}_l(\mathbf{E}_{\text{RADIO}}(\mathbf{I})_l)\big)\in\mathbb{R}^{h\times w\times1024}$. Each $1\times1$ convolutional block $\mathcal{C}_l$ compresses its map to 512 channels; learned softmax-normalized weights $\alpha_l$ combine the maps, and a $3\times3$ convolutional block $\mathcal{R}$ refines the sum to 1024 channels.



We then lift the dense image features onto the partial point cloud. For each point $\mathbf{P}_i$, we gather $\mathbf{f}_i=\mathbf{F}_{\text{rgb}}[\lfloor\pi(\mathbf{P}_i)/16\rfloor]$ at its patch-grid index, where $\pi(\cdot)$ gives crop-pixel coordinates and 16 is the patch size. Following the point-wise fusion spirit of DenseFusion~\cite{wang2019densefusion}, we concatenate each 3D point with its lifted visual feature (denoted by $\odot$) and process the fused point set with a Dynamic Graph Convolutional Neural Network (DGCNN) encoder $\mathbf{E}_{\text{GCN}}(\cdot)$. This yields the fused embedding $\mathbf{F}_{\text{fuse}} = \mathbf{E}_{\text{GCN}}(\{\mathbf{P}_i \odot \mathbf{f}_i\}_{i=1}^{N}) \in \mathbb{R}^{n \times d}$, which serves as local geometry-aware input tokens for the Transformer encoder, where $n = 128$ and $d = 512$ denote the number of tokens and the embedding dimension, respectively.

\subsection{MoE-Based Object Encoding}

The DGCNN tokens encode local surface neighborhoods and point-wise visual cues. We process these tokens with a stack of Transformer blocks with multi-head self-attention~\cite{vaswani2017attention}, following the use of Transformers for point cloud completion and geometry modeling~\cite{yu2021pointr,yu2023adapointrdiversepointcloud}. This stage integrates local evidence across the object crop to form an object-level representation for metric prediction.

To increase representation capacity with conditional expert aggregation, we replace the standard feed-forward layers in each Transformer block with Mixture-of-Experts (MoE) layers~\cite{du2022glam,fedus2022switch,liu2024deepseek}. Each MoE layer contains an independent set of experts and uses a linear gating network to compute softmax probabilities over $N_e=8$ experts. For each token, we select the top-2 experts according to the gating scores, route the token to these experts, and aggregate their outputs using the selected probabilities renormalized to sum to one. A soft load-balancing penalty encourages balanced gate probabilities during training~\cite{shazeer2017outrageously}. Finally, the encoded tokens are aggregated by global pooling to produce a compact object representation for direct 6D pose, size, and shape prediction.


\subsection{Direct Pose-Size and Shape Decoding}

F2G-Pose uses two lightweight decoding branches on top of the shared object representation. The pose-size branch predicts the metric state directly, while the shape branch predicts complete geometry in the camera frame. Both branches are trained jointly, so dense geometric supervision can regularize the representation used for direct pose and size prediction without adding any test-time optimization or candidate extraction.

\noindent\textbf{Pose and Size Estimation.}
The pose-size branch regresses rotation, translation, and metric size directly from the global object representation. For rotation, we adopt the continuous 6D representation~\cite{zhou2019continuity}, which uses the first two columns of the rotation matrix $\mathbf{R} \in \text{SO}(3)$ and reconstructs a valid matrix via orthogonalization. This parameterization improves optimization stability while keeping the prediction deterministic and feed-forward. The branch outputs the final pose-size estimate directly, without sampling pose candidates or applying post-hoc aggregation.


\noindent\textbf{Camera-Frame Shape Completion.}
From the extracted global feature vector, we first regress a coarse complete object shape $\mathbf{P}_{\text{coarse}} \in \mathbb{R}^{512 \times 3}$ using a lightweight MLP. Since the input partial point cloud $\mathbf{P}$ contains observed metric geometry, we concatenate $\mathbf{P}_{\text{coarse}}$ with farthest-point-sampled observations from $\mathbf{P}$ to form a combined point set, which is then processed through another MLP with a Sigmoid activation to produce a confidence score for each point. This selects reliable points for fusion, producing the refined fused point cloud $\mathbf{P}_{\text{fuse}} \in \mathbb{R}^{512 \times 3}$. Finally, each point in $\mathbf{P}_{\text{fuse}}$ is augmented with the global feature vector and passed through an MLP to predict four local offsets per point, yielding $\mathbf{P}_\text{dense} \in \mathbb{R}^{2048 \times 3}$. The shape is reconstructed directly in the camera frame, providing a geometry output and a dense auxiliary signal for learning metric-consistent object representations.



\subsection{Training Objectives}
\noindent\textbf{Pose and Size Regression Loss.}
We supervise pose and size with Smooth L1 losses. For rotation, we account for object symmetries following the benchmark treatment in~\cite{zhang2025omni6dpose}, so symmetry-equivalent rotations are treated as valid targets rather than arbitrary errors. Specifically, for each axis, an object's symmetry is categorized as no symmetry, 90-degree rotational symmetry, 180-degree rotational symmetry, or arbitrary-angle rotational symmetry. Based on this classification, we generate a set of valid ground-truth rotation matrices and compare the predicted rotation against all candidates. The loss is computed with respect to the closest ground-truth rotation.
The pose and size estimation losses are defined as:
\begin{align}
    \mathcal{L}_\text{rot} &= \min_{\mathbf{R}^\text{gt}_i \in \mathcal{G}_\mathbf{R}} \ \text{SmoothL1}(\hat{\mathbf{r}}, \, \mathbf{r}^\text{gt}_i), \\
    \mathcal{L}_\text{trans} &= \text{SmoothL1}(\mathbf{t}, \ \mathbf{t}^\text{gt}), \\
    \mathcal{L}_\text{size} &= \text{SmoothL1}(\mathbf{s}, \ \mathbf{s}^\text{gt}).
\end{align}
Here, $\hat{\mathbf{r}}\in\mathbb{R}^6$ is the rotation head's raw output, and $\mathbf{r}^\text{gt}_i$ stacks the first two columns of $\mathbf{R}^\text{gt}_i$. The translation and size targets are $\mathbf{t}^\text{gt}$ and $\mathbf{s}^\text{gt}$. The set $\mathcal{G}_\mathbf{R}=\{\mathbf{R}^\text{gt}_i\}_{i=1}^{M}$ contains symmetry-equivalent target rotations; the minimum is taken over their 6D representations before orthogonalization.

\noindent\textbf{Reconstruction Loss.}
For camera-frame shape completion, we adopt the Chamfer Distance with L1 norm, following~\cite{yu2021pointr}, to supervise both the coarse and dense point cloud outputs. Specifically, we define two reconstruction losses: one for the selected coarse point cloud $\mathbf{P}_\text{fuse}$ and another for the final dense reconstruction $\mathbf{P}_\text{dense}$. The Chamfer Distance measures the average closest-point distance between the predicted and ground-truth point sets, encouraging accurate geometry at different stages of the pipeline.

The reconstruction losses are formulated as follows:
\begin{align}
    \mathcal{L}_\text{recon} = \text{Chamfer}(\mathbf{P}_\text{fuse}, \mathbf{P}^\text{gt}) + \text{Chamfer}(\mathbf{P}_\text{dense}, \mathbf{P}^\text{gt}).
\end{align}
Here, $\mathbf{P}^\text{gt}$ is the same ground truth complete point cloud for both terms. These losses jointly guide the model to generate increasingly accurate shapes from coarse to fine resolution.

\noindent\textbf{MoE Balancing Loss.}
To encourage balanced expert usage in the MoE layers, we incorporate a load-balancing auxiliary loss, following the Switch Transformer design~\cite{fedus2022switch}. For each MoE layer with $N_e$ experts, the gating network produces a probability distribution over the experts. The auxiliary loss $\mathcal{L}_{\text{aux}}$ is defined to encourage a uniform distribution of tokens across experts:
\begin{equation}
\mathcal{L}_{\text{aux}} = \frac{1}{L} \sum_{l=1}^{L} \left( N_e \cdot \sum_{i=1}^{N_e} f_i^{(l)} \cdot P_i^{(l)} \right),
\end{equation}
Here, $L$ is the number of MoE layers and $P_i^{(l)}$ is the mean gating probability over all tokens for expert $i$ in layer $l$. We use the soft variant $f_i^{(l)}=P_i^{(l)}$, so each layer contributes $N_e\sum_i(P_i^{(l)})^2$.

\noindent\textbf{All Losses.}
Our final objective integrates direct pose-size prediction, camera-frame shape completion, and MoE load balancing:
\begin{equation}
\mathcal{L} = \mathcal{L}_{\text{recon}} + \mathcal{L}_{\text{rot}} + \mathcal{L}_{\text{trans}} + \mathcal{L}_{\text{size}} + \lambda \mathcal{L}_{\text{aux}},
\end{equation}
where $\lambda$ is a balancing coefficient. In practice, we set $\lambda = 0.01$.




\section{Experiments}

\noindent\textbf{Datasets.} We use SOPE~\cite{zhang2025omni6dpose} as the training source for F2G-Pose; it contains 5K synthetic object instances from 149 categories with category-level canonical pose, metric size, and complete shape supervision.
Pose-size evaluation uses SOPE and ROPE~\cite{zhang2025omni6dpose}: SOPE measures synthetic benchmark performance, while ROPE evaluates transfer to scanned real objects under the same category taxonomy.
For camera-frame geometry evaluation, we use SOPE and HouseCat6D~\cite{jung2024housecat6d}, where complete object geometry is available.
For additional real-world transfer, we evaluate selected HANDAL objects~\cite{guo2023handal}, including hammers, mugs, pans, and wrench-like tools. The 92 instances from seven HANDAL categories map to six SOPE categories; selection follows taxonomy correspondence. Thus, ROPE and this HANDAL subset evaluate seen-category, unseen-instance transfer (Appendix~\ref{app:handal_categories}).

\noindent\textbf{Baselines and metrics.} For category-level pose-size estimation, we compare with NOCS, SGPA, IST-Net, HS-Pose, GenPose, and GenPose++ using the SOPE/ROPE benchmark results reported by Omni6DPose~\cite{zhang2025omni6dpose}.
Joint-RegDiff~\cite{lee2025jointlearningposeregression} and RFMPose~\cite{ouyang2025rfmpose} are report-only comparisons from their papers, using public metrics available under the SOPE/ROPE taxonomy.
For reference-based comparisons on real datasets, we additionally report FoundationPose~\cite{wen2024foundationpose} and Any6D~\cite{lee2025any6d}.
For runtime comparison, we benchmark executable methods under a unified inference protocol using the same segmentation masks, RGB-D crops, depth preprocessing, batch size 1, and GPU-synchronized timing on one RTX 4090; accuracy results follow the evaluation sources specified above.
Pose-size accuracy is evaluated with AUC of 3D bounding-box IoU at thresholds 25/50/75, VUS under joint rotation-translation thresholds, and mean rotation/translation error.
Shape completion is evaluated using Chamfer-L1 distance.

\subsection{Comparison with State-of-the-Art Methods}
\begin{table}[t]
\caption{{Category-level pose-size estimation on ROPE and SOPE under the shared SOPE/ROPE category taxonomy.
`--' indicates unavailable or non-hardware-normalized metrics. Joint-RegDiff and RFMPose are report-only comparisons from their papers.
\textsuperscript{*} indicates our pose-only inference mode (shape head disabled); the full model runs at 23.7 FPS including shape completion.} \label{table:rope_sope}}
\centering
\footnotesize
\setlength\tabcolsep{2pt}
\renewcommand{\arraystretch}{1.0}
\resizebox{\linewidth}{!}{
\begin{tabular}{c|c|ccc|cccc|cc|c}
\toprule
\multirow{2}{*}{Dataset} & \multirow{2}{*}{Method}
& \multicolumn{3}{c|}{AUC $\uparrow$}
& \multicolumn{4}{c|}{VUS $\uparrow$}
& \multicolumn{2}{c|}{Mean error $\downarrow$}
& \multirow{2}{*}{FPS $\uparrow$} \\
&
& IoU25 & IoU50 & IoU75
& 5$^\circ$2cm & 5$^\circ$5cm & 10$^\circ$2cm & 10$^\circ$5cm
& rot($^\circ$) & trans(cm)
&  \\
\midrule
\multirow{8}{*}{ROPE}
& NOCS~\cite{wang2019normalized}      & 0.0 & 0.0 & 0.0 & 0.0 & 0.0 & 0.0 & 0.0 & -- & -- & 5 \\
& SGPA~\cite{chen2021sgpa}       & 10.5 & 2.0 & 0.0 & 4.3 & 6.7 & 9.3 & 15.0 & 60.2 & 2.81 & 6 \\
& IST-Net~\cite{liu2023net}   & 28.7 & 10.6 & 0.5 & 2.0 & 3.4 & 5.3 & 8.8 & 78.4 & 3.40 & 35 \\
& HS-Pose~\cite{zheng2023hs}   & 31.6 & 13.6 & 1.1 & 3.5 & 5.3 & 8.4 & 12.7 & 63.3 & 3.02 & \textbf{50} \\
& GenPose~\cite{zhang2023genpose}   & --   & --   & --  & 6.6 & 9.6 & 13.1 & 19.3 & 42.2 & 2.05 & 6 \\
& GenPose++~\cite{zhang2025omni6dpose} & 39.0 & 19.1 & 2.0 & 10.0 & 15.1 & 19.5 & 29.4 & 30.7 & 1.28 & 4 \\
& Joint-RegDiff~\cite{lee2025jointlearningposeregression} & --   & --   & --  & 10.9 & -- & 20.8 & -- & -- & -- & -- \\
& RFMPose~\cite{ouyang2025rfmpose} & 42.1 & 21.0 & 2.2 & 10.4 & 15.7 & 21.0 & 30.8 & -- & -- & -- \\
\cmidrule{2-12}
& \textbf{Ours}       & \textbf{46.5} & \textbf{27.8} & \textbf{5.8} & \textbf{11.3} & \textbf{15.9} & \textbf{21.8} & \textbf{31.1} & \textbf{27.5} & \textbf{1.19} & 28$^{*}$ \\
\midrule[0.5pt] \multirow{7}{*}{SOPE} & NOCS~\cite{wang2019normalized} & 0.0 & 0.0 & 0.0 & 0.0 & 0.0 & 0.0 & 0.0 & -- & -- & 5 \\ & SGPA~\cite{chen2021sgpa} & 13.3 & 3.2 & 0.0 & 7.7 & 10.1 & 15.0 & 20.4 & 33.8 & 2.09 & 6 \\ & IST-Net~\cite{liu2023net} & 36.5 & 16.9 & 1.4 & 3.6 & 5.1 & 8.6 & 11.4 & 60.4 & 3.72 & 35 \\ & HS-Pose~\cite{zheng2023hs} & 40.1 & 21.7 & 3.2 & 6.3 & 8.0 & 13.6 & 17.3 & 39.9 & 2.46 & \textbf{50} \\ & GenPose~\cite{zhang2023genpose} & -- & -- & -- & 11.9 & 14.4 & 21.2 & 26.3 & 26.1 & 1.62 & 6 \\ & GenPose++~\cite{zhang2025omni6dpose} & 50.1 & 31.9 & 6.4 & 18.4 & 23.0 & 31.9 & 40.2 & 19.9 & 1.14 & 4 \\ \cmidrule{2-12} & \textbf{Ours} & \textbf{59.6} & \textbf{43.7} & \textbf{16.5} & \textbf{20.5} & \textbf{24.2} & \textbf{35.2} & \textbf{42.1} & \textbf{15.7} & \textbf{0.88} & 28$^{*}$ \\
\bottomrule
\end{tabular}
}
\end{table}

\noindent\textbf{Pose-size accuracy and latency.} Table~\ref{table:rope_sope} shows that F2G-Pose improves the accuracy-latency operating point of category-level RGB-D pose estimation.
On SOPE, F2G-Pose improves GenPose++ from 31.9 to 43.7 AUC at IoU50 and reduces mean rotation error from $19.9^\circ$ to $15.7^\circ$; on ROPE, it improves AUC across all IoU thresholds and matches or exceeds the available high-precision VUS results.
This gain is not only a speedup from removing iterative sampling.
The larger improvement on IoU and rotation-sensitive metrics indicates that direct prediction benefits from stronger conditioning: dense image features preserve category and part cues, local point neighborhoods preserve observed surface evidence, and global token interactions integrate these cues into a metric pose-size estimate.
At the same time, F2G-Pose runs at 28 FPS in pose-only mode and 24 FPS in full-model mode, compared with 4 FPS for GenPose++, yielding a single-pass estimate without candidate scoring, filtering, or clustering. More qualitative examples are provided in Appendix~\ref{app:qualitative}.

\begin{figure*}[t]
    \centering
    \includegraphics[width=\textwidth]{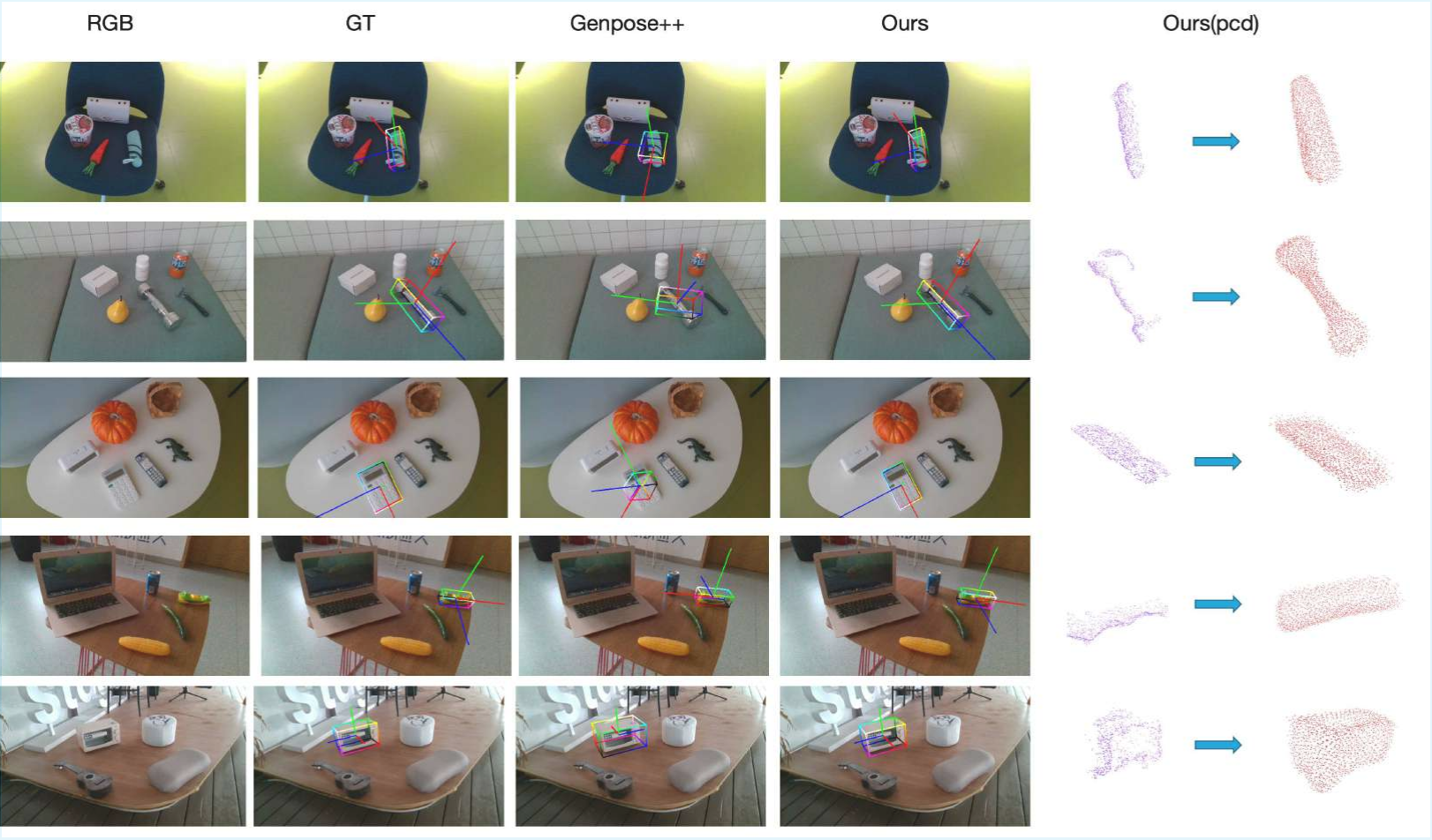}
    \caption{Qualitative results on ROPE. We show the input RGB image, ground-truth pose, poses from GenPose++ and F2G-Pose, and the predicted camera-frame shape.}
    \label{fig:examples}
\end{figure*}

\begin{table}[tbp]
\centering
\caption{GenPose++ candidate diagnostics on ROPE. The sampled set can contain accurate hypotheses, but deployed accuracy depends on candidate extraction.}
\label{tab:genpose_candidate}
\small
\setlength\tabcolsep{4pt}
\renewcommand{\arraystretch}{0.95}
\begin{tabular}{l|cccc}
\toprule
Analysis & \multicolumn{4}{c}{Rotation Error (degrees)} \\
\cmidrule(lr){2-5}                                              
& Mean $\downarrow$ & Median $\downarrow$ & $>45^\circ$ $\downarrow$ & $\leq 5^\circ$ $\uparrow$ \\
\midrule
All candidates    & 39.3 & 8.2  & 26.9\% & 34.0\% \\
Best-of-50 oracle & 9.0  & 1.78 & 6.1\%  & 79.3\% \\
Final pipeline    & 30.7 & 5.2  & 19.1\% & 48.3\% \\
\bottomrule
\end{tabular}
\end{table}

\noindent\textbf{Candidate-to-decision and time-budget analysis.}
Table~\ref{tab:genpose_candidate} examines the candidate behavior of GenPose++ as a strong generative baseline. Its candidate distribution is heavy-tailed: although the median rotation error is only $8^\circ$, the mean rises to $39^\circ$, with 26.9\% of sampled candidates exceeding $45^\circ$. With $K=50$ samples, an oracle selects a candidate within $5^\circ$ for 79.3\% of instances, achieving a median error of $1.78^\circ$. However, a deployed pipeline must still score, filter, cluster, or aggregate these candidates, so its final prediction can be substantially worse than the oracle. This highlights the candidate-to-decision gap discussed in the introduction: candidate-set quality and deployed prediction quality are related but not equivalent. Candidate quality also varies with object structure; no-symmetry objects have the largest mean rotation error, suggesting that sampling difficulty is driven not only by pose ambiguity but also by distinctive yet partially observed geometry. Additional candidate diagnostics by symmetry are provided in Appendix~\ref{app:candidate_diagnostics}.


\begin{figure}[tbp]

    \centering
    \includegraphics[width=0.90\linewidth]{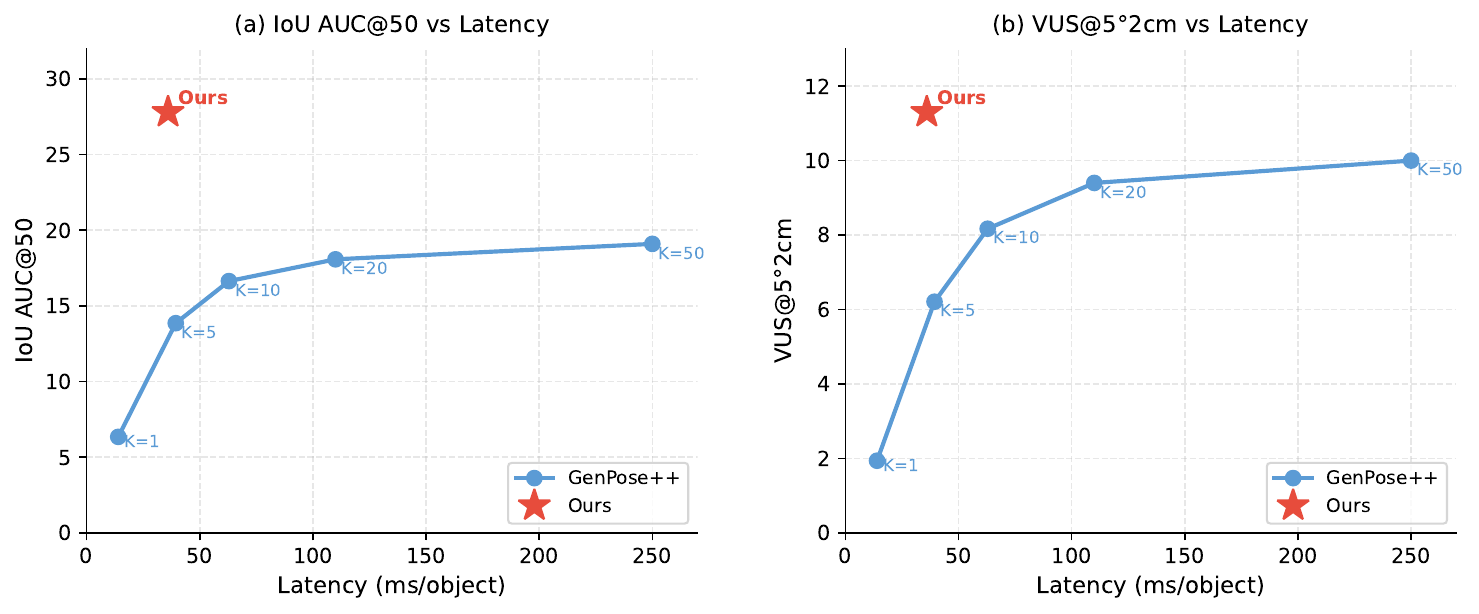}
    \caption{Pareto analysis of GenPose++ candidate budget on ROPE.}
    \label{fig:pareto_front}

\end{figure}

Figure~\ref{fig:pareto_front} evaluates the same issue under a real-time budget.
Reducing diffusion candidates improves responsiveness but lowers final accuracy because candidate extraction becomes less reliable.
F2G-Pose avoids this trade-off by optimizing the final pose-size decision directly.


\begin{table}[t]
\centering
\begin{minipage}[t]{0.40\linewidth}
\centering
\captionsetup{font=scriptsize,skip=1pt}
\caption{Camera-frame shape reconstruction on SOPE and HouseCat6D from observed partial object geometry.}
\label{table:cdl1_rope_sope}
\tiny
\setlength\tabcolsep{2.5pt}
\renewcommand{\arraystretch}{0.86}
\resizebox{0.90\linewidth}{!}{
\begin{tabular}{l|cc}
\toprule
\multirow{2}{*}{Method}
    & \multicolumn{2}{c}{Chamfer-L1 ($\times 10^{-3}$) $\downarrow$} \\
\cmidrule(lr){2-3}
    & SOPE & HouseCat6D \\
\midrule
FoldingNet & 62.72 & 73.2 \\
Pointr & 29.87 & 40.2 \\
AdaPointr & 24.41 & 32.3 \\
\textbf{Ours} & \textbf{5.42} & \textbf{7.1} \\
\bottomrule
\end{tabular}}
\end{minipage}
\hfill
\begin{minipage}[t]{0.57\linewidth}
\centering
\caption{Component ablation on ROPE.}
\label{table:ablation}
\scriptsize
\setlength\tabcolsep{1.5pt}
\renewcommand{\arraystretch}{0.94}
\resizebox{\linewidth}{!}{
\begin{tabular}{l|ccc|cc|c}
\toprule
\multirow{2}{*}{Variant}
& \multicolumn{3}{c|}{AUC $\uparrow$}
& \multicolumn{2}{c|}{VUS $\uparrow$}
& \multirow{2}{*}{FPS $\uparrow$} \\
\cmidrule(lr){2-4} \cmidrule(lr){5-6}
& IoU25 & IoU50 & IoU75 & 5$^\circ$5cm & 10$^\circ$2cm & \\
\midrule
\textbf{Full}        & \textbf{46.5} & \textbf{27.8} & \textbf{5.8} & \textbf{15.9} & \textbf{21.8} & 23.7 \\
w/o RGB              & 34.5 & 16.2 & 2.2 &  6.5 & 13.2 & \textbf{29.5} \\
w/o MoE              & 41.0 & 24.0 & 3.9 &  8.7 & 17.8 & 19.2 \\
w/o Shape Sup.       & 39.5 & 23.2 & 3.5 &  8.1 & 16.5 & 27.8 \\
\bottomrule
\end{tabular}}
\end{minipage}
\vspace{-0.08in}
\end{table}

\noindent\textbf{Geometry output and shape completion.} Table~\ref{table:cdl1_rope_sope} evaluates camera-frame geometry output.
The point-completion baselines are trained on the SOPE training split with the same partial-to-complete camera-frame targets.
Compared with these geometry-only baselines~\cite{yang2018foldingnetpointcloudautoencoder,yu2021pointr,yu2023adapointrdiversepointcloud}, F2G-Pose achieves substantially lower Chamfer-L1 error on both SOPE and HouseCat6D. The qualitative results in Fig.~\ref{fig:examples} show that F2G-Pose successfully reconstructs complete object geometries.
Unlike these geometry-only completion baselines, F2G-Pose predicts shape from a shared RGB-D representation: depth points anchor the visible metric surface, and lifted RGB foundation features provide category, texture, and part cues for completing occluded regions.
Since the completed geometry is predicted directly in the camera frame, it can also be consumed by simple downstream planning modules without an additional canonical-to-camera alignment step. Because the inputs and supervision differ across systems and camera-frame CD also reflects global alignment, Appendix~\ref{app:controlled_completion} provides an additional matched RGB-D canonical-frame comparison.

\subsection{Ablation Studies}
\noindent\textbf{Component ablation.}
Table~\ref{table:ablation} isolates the main design factors of F2G-Pose on ROPE.
Removing RGB foundation features leads to the largest performance drop, particularly on the rotation-sensitive VUS metrics, showing that point-wise visual lifting provides pose-relevant appearance and semantic cues beyond partial geometry alone.
Replacing the MoE Transformer with a parameter-equivalent dense FFN also reduces accuracy, suggesting that conditional capacity helps model category-level object diversity.
Removing auxiliary shape supervision further degrades pose-size prediction, indicating that dense camera-frame geometry supervision regularizes the shared RGB-D representation and directly benefits the final pose-size estimate.

\begin{table}[t]
\centering
\begin{minipage}[t]{0.50\linewidth}
\centering
\caption{Representation and backbone study on ROPE. Representation variants use RADIO as the default backbone.}
\label{tab:representation_study}
\scriptsize
\setlength\tabcolsep{1.4pt}
\renewcommand{\arraystretch}{0.86}
\resizebox{\linewidth}{!}{
\begin{tabular}{l|ccc}
\toprule
Variant & AUC$_{\mathrm{IoU25}}$ & AUC$_{\mathrm{IoU50}}$ & VUS$_{5^\circ5\mathrm{cm}}$ \\
\midrule
\multicolumn{4}{l}{\textit{Representation variants}} \\
Global PointNet & 35.2 & 17.8 & 6.3 \\
DGCNN tokens & 38.6 & 21.6 & 7.8 \\
DGCNN + Trans. & 41.0 & 24.0 & 8.7 \\
DGCNN + MoE w/o Shape & 39.5 & 23.2 & 8.1 \\
\midrule
\multicolumn{4}{l}{\textit{Backbone variants}} \\
Full + DINOv2 & 45.5 & 27.0 & 13.7 \\
Full + CLIP & 44.8 & 26.3 & 12.5 \\
\midrule
\textbf{Full} & \textbf{46.5} & \textbf{27.8} & \textbf{15.9} \\
\bottomrule
\end{tabular}}
\end{minipage}
\hfill
\begin{minipage}[t]{0.48\linewidth}
\centering
\caption{MoE routing and capacity analysis on ROPE. Routing entropy is normalized by $\log 8$; higher is more balanced.}
\label{tab:moe_routing}
\scriptsize
\setlength\tabcolsep{1.2pt}
\renewcommand{\arraystretch}{1.22}
\resizebox{\linewidth}{!}{
\begin{tabular}{l|ccccc}
\toprule
FFN & IoU50 & VUS & Ent. & Load & FPS \\
\midrule
Dense FFN & 24.0 & 8.7 & -- & -- & 19.2 \\
MoE top-1 & 25.6 & 13.5 & 0.89 & 24\% & \textbf{26.1} \\
MoE top-2, no balance & 26.0 & 14.2 & 0.71 & 39\% & 23.9 \\
\textbf{MoE top-2} & \textbf{27.8} & \textbf{15.9} & 0.95 & 18\% & 23.7 \\
MoE top-4 & 28.0 & 16.1 & \textbf{0.96} & \textbf{17\%} & 18.5 \\
\bottomrule
\end{tabular}}
\end{minipage}
\vspace{-0.08in}
\end{table}

\noindent\textbf{Representation analysis.}
Table~\ref{tab:representation_study} studies how different RGB-D representations affect direct pose-size prediction.
All variants are trained with the same pose-size losses, while varying the feature aggregation strategy, contextual encoder, foundation backbone, and the use of auxiliary shape supervision.
Replacing global feature pooling with DGCNN tokens improves both IoU and rotation-sensitive VUS, showing that local surface neighborhoods provide important geometric cues for direct pose estimation.
Adding Transformer context further improves performance by integrating local RGB-D evidence into an object-level representation.
The full model achieves the best results, indicating that local geometric tokenization, object-level context, and camera-frame geometry supervision are complementary.
Finally, replacing RADIO with DINOv2 or CLIP causes only modest degradation in IoU50, suggesting that the improvement is driven primarily by the architecture and supervision rather than by a specific foundation backbone.
Additional backbone, routing, and shape-decoder ablations are summarized in Appendix~\ref{app:ablation}.



\noindent\textbf{MoE routing analysis.} Table~\ref{tab:moe_routing} shows that top-2 MoE provides a favorable accuracy-latency trade-off.
Top-1 routing is faster but less accurate, while top-4 brings only marginal accuracy gains at a higher latency cost.
Compared with top-2 routing without balancing, the load-balanced top-2 MoE achieves higher routing entropy, lower load concentration, and better pose accuracy.
These results suggest that balanced conditional capacity improves the shared RGB-D representation while keeping inference cost moderate. Further instance-level analysis finds overlapping category-related expert preferences (Appendix~\ref{app:moe_specialization}).

\begin{table}[tbp]
\centering

\caption{Performance on reflective and transparent materials across ROPE and HouseCat6D.}
\label{tab:mat_performance}
\small
\setlength\tabcolsep{3pt}
\renewcommand{\arraystretch}{1.25}
\resizebox{\linewidth}{!}{
\begin{tabular}{l|ccccc|c}
\toprule
\multirow{2}{*}{Category}
& \multicolumn{5}{c|}{ROPE}
& HouseCat6D \\
\cmidrule(lr){2-6} \cmidrule(lr){7-7}
& AUC$_{\mathrm{IoU25}}$ $\uparrow$
& AUC$_{\mathrm{IoU75}}$ $\uparrow$
& VUS$_{5^\circ2\mathrm{cm}}$ $\uparrow$
& Rot. Err. $\downarrow$
& Trans. Err. $\downarrow$
& CD $\downarrow$ \\
\midrule
Transparent & 43.5 & 4.7 & 10.4 & 25.3 & 1.19 & 6.2 \\
Reflective  & 42.6 & 4.5 & 9.8  & 25.2 & 1.25 & 10.1 \\
Average     & 46.5 & 5.8 & 11.3 & 27.5 & 1.19 & 7.1 \\
\bottomrule
\end{tabular}
}
\end{table}

\noindent\textbf{Robustness under challenging materials.} Table~\ref{tab:mat_performance} reports performance on transparent and reflective objects, which often corrupt depth observations.
Compared with the overall ROPE performance, F2G-Pose shows moderate degradation on these material subsets, with reflective objects being more difficult than transparent ones, especially in VUS and Chamfer distance.
The performance drop is most apparent in IoU75, where depth holes, specular artifacts, and missing thin structures can perturb the estimated object extent.
A similar trend is observed on HouseCat6D, where reconstruction remains accurate for many transparent objects because their shapes are regular despite noisy depth.
Overall, F2G-Pose remains reasonably robust under material-induced sensing noise, while precise box alignment is still sensitive to imperfect depth observations.
Controlled RGB and depth perturbation results are reported in Appendix~\ref{app:controlled_robustness}, with additional material examples in Fig.~\ref{fig:special_case}. SAM-mask and controlled erosion/dilation results are provided in Appendix~\ref{app:mask_sensitivity}.


\vspace{-0.06in}
\subsection{Real-World Evaluation and Deployment}

\noindent\textbf{Cross-dataset real-world evaluation.}

\begin{wraptable}{r}{0.44\linewidth}
\centering
\vspace{-0.36in}
\captionsetup{font=scriptsize,skip=1pt}
\caption{Cross-dataset ROPE/HANDAL evaluation.}
\label{table:rope_handal_auc}
\tiny
\setlength\tabcolsep{2.2pt}
\renewcommand{\arraystretch}{0.80}
\begin{tabular}{@{}ll|ccc@{}}
\toprule
\multirow{2}{*}{Data} & \multirow{2}{*}{Method} & \multicolumn{3}{c}{AUC $\uparrow$} \\
& & IoU25 & IoU50 & IoU75 \\
\midrule
\multirow{4}{*}{ROPE}
& FoundationPose (1 ref) & 35.0 & 18.0 & 3.1 \\
& Any6D (1 ref)     & 37.2 & 19.4 & 3.5 \\
& Any6D (0 ref)     & 22.5 & 8.1  & 0.3 \\
& \textbf{Ours}     & \textbf{46.5} & \textbf{27.8} & \textbf{5.8} \\
\midrule[0.5pt]
\multirow{3}{*}{HANDAL}
& Any6D (0 ref)     & 14.5 & 3.8 & 0.0 \\
& GenPose++ (0 ref) & 16.7 & 4.3 & 0.1 \\
& \textbf{Ours}     & \textbf{35.0} & \textbf{11.7} & \textbf{0.3} \\
\bottomrule
\end{tabular}
\end{wraptable}

Table~\ref{table:rope_handal_auc} evaluates cross-dataset transfer on real-world manipulable objects captured under different setups. The zero-/one-reference baseline protocols are detailed in Appendix~\ref{app:reference_protocols}.
On HANDAL, F2G-Pose outperforms the reference-free methods Any6D and GenPose++, while on ROPE it further surpasses the single-reference method FoundationPose and the reference-free method Any6D, without requiring an instance reference.
These results complement the controlled SOPE/ROPE benchmark by exposing the model to additional real objects, masks, backgrounds, and sensor noise patterns.
Despite these transfer gains, the low IoU75 values show that high-precision box alignment remains challenging under cross-dataset sensing and annotation differences.
These results provide additional evidence of F2G-Pose's robustness under real-world distribution shifts, complementing the controlled category-level analysis on SOPE/ROPE; qualitative HANDAL examples are shown in Fig.~\ref{fig:handal_cases}.


\noindent\textbf{Robot manipulation.} We validate the final pose-size decision in a real pick-and-place setup using an ARX robotic arm, a parallel-jaw gripper, and an external RealSense RGB-D camera.
All methods receive the same object masks and RGB-D crops, and the evaluated objects include bottles, bowls, cans, mugs, cups, hammers, pans, and wrench-like tools.
The predicted pose and metric size determine the grasp approach and placement target; the completed camera-frame shape is used only for collision checking.
Each trial randomizes object placement within a tabletop start region, and success requires grasping, transporting, and placing the object at the target without dropping or collision. The full model achieves 74/80 successful picks and 68/80 successful placements.

\begin{figure}[tbp]
\centering
\begin{minipage}[t]{0.49\linewidth}
\vspace{0pt}
\captionsetup{font=footnotesize,skip=3pt}
\centering

\captionof{table}{Real-world pose-conditioned pick-and-place on ARX (80 trials per method). Lat. is mean perception latency; Ready@50 is the measured percentage of usable decisions returned within 50\,ms; Time is mean pick-and-place duration over successful trials.}
\label{tab:robot_pick_place}
\scriptsize
\setlength\tabcolsep{1.2pt}
\renewcommand{\arraystretch}{1.0}
\resizebox{\linewidth}{!}{
\begin{tabular}{p{0.30\linewidth}ccccc}
\toprule
\raggedright Method & \shortstack{Lat.\\(ms)} & Ready@50 & Pick & Place & \shortstack{Time\\(s)} \\
\midrule
\raggedright GenPose++ full~\cite{zhang2025omni6dpose} & 256 & 0\%   & 66/80 & 57/80 & 9.8 \\
\raggedright GenPose++ time-limited~\cite{zhang2025omni6dpose} & 49 & 100\% & 56/80 & 45/80 & 7.6 \\
\raggedright Ours depth-only & 31 & 100\% & 61/80 & 50/80 & 6.2 \\
\raggedright Ours w/o shape completion & 29 & 100\% & 68/80 & 57/80 & \textbf{6.0} \\
\raggedright \textbf{Ours full} & 36 & 100\% & \textbf{74/80} & \textbf{68/80} & 6.1 \\
\bottomrule
\end{tabular}}
\end{minipage}
\hfill
\begin{minipage}[t]{0.49\linewidth}
\vspace{0pt}
\captionsetup{font=footnotesize,skip=3pt}
\centering

\includegraphics[width=\linewidth]{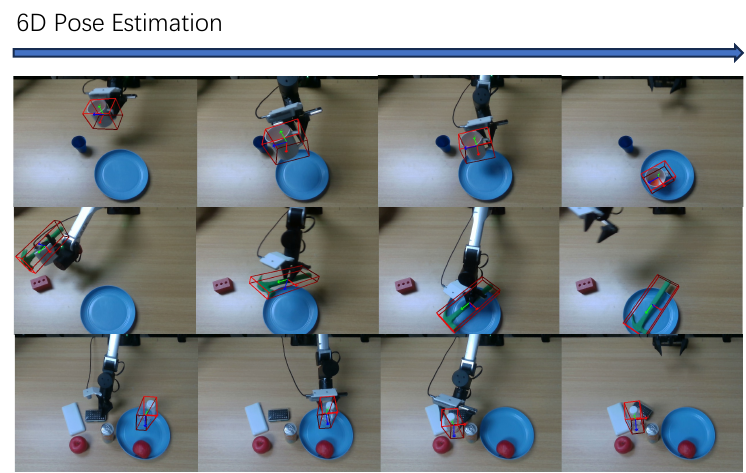}
\captionof{figure}{Real-world ARX manipulation setup and example pick-and-place trials.}
\label{fig:real_manip}
\end{minipage}
\end{figure}

Figure~\ref{fig:real_manip} illustrates the physical setup and representative trials, while Table~\ref{tab:robot_pick_place} shows that F2G-Pose preserves high manipulation success while meeting a real-time perception budget.
GenPose++ is effective when run with its full sampling and clustering pipeline, but the full pipeline is too slow for the 50 ms control budget; reducing candidates improves responsiveness at the cost of lower pick-and-place success.
The depth-only and w/o-shape variants further show that dense visual lifting and camera-frame shape supervision improve deployed pick-and-place success.

\noindent\textbf{Failure analysis.} In qualitative benchmark cases, failures mainly occur when corrupted depth removes thin or asymmetric parts, severe occlusion hides functional geometry, or symmetric low-texture objects admit 90$^\circ$/180$^\circ$ alternatives.
Additional examples are shown in Appendix~\ref{app:failure_modes}.

\section{Conclusion}

We present F2G-Pose, a direct RGB-D framework that estimates category-level 6D pose, metric size, and camera-frame shape from a single observation.
By lifting dense visual foundation features to partial point clouds and encoding local-global geometry with a MoE Transformer, F2G-Pose predicts the deployed pose-size state in one feed-forward pass.
Experiments on SOPE, ROPE, HouseCat6D, HANDAL, and ARX manipulation show a favorable balance of accuracy, inference latency, and manipulation success, while the shape branch provides dense geometric supervision and approximate planning geometry.

\noindent\textbf{Limitations.} 
F2G-Pose assumes a segmented RGB-D crop at inference time, so mask failures and severe depth corruption can propagate to pose, size, and shape. 
The method remains sensitive to thin or asymmetric parts, severe occlusion, and symmetric low-texture objects that admit multiple plausible rotations. 
High-precision metric alignment remains weak under cross-dataset domain shift, which may require downstream refinement for tasks with tight geometric tolerances. Symmetry-aware supervision does not eliminate ambiguity under severe occlusion or near-symmetry, and the model provides no calibrated uncertainty or abstention mechanism.
Future work will address mask-free operation, broader real-object coverage, articulated or deformable categories, and active perception for heavy occlusion.

\noindent\textbf{Broader Impact.}
F2G-Pose advances category-level object perception for robotics and augmented reality. While improved pose estimation can benefit assistive manipulation and human--robot collaboration, it could also be applied in surveillance or autonomous systems without adequate safety oversight. We encourage responsible deployment with appropriate human supervision.

\clearpage
\begin{ack}
This work received no third-party funding or external support. The authors declare no competing interests.
\end{ack}

\bibliographystyle{plainnat}
\bibliography{ref}

@String(CVPR  = {IEEE Conf. Comput. Vis. Pattern Recog.})

@String(ICCV  = {Int. Conf. Comput. Vis.})

@String(CVPR  = {CVPR})

@String(ICCV  = {ICCV})

@inproceedings{wang2019densefusion,
  title={{DenseFusion}: {6D} object pose estimation by iterative dense fusion},
  author={Wang, Chen and Xu, Danfei and Zhu, Yuke and Mart{\'\i}n-Mart{\'\i}n, Roberto and Lu, Cewu and Fei-Fei, Li and Savarese, Silvio},
  booktitle={Proceedings of the IEEE/CVF conference on computer vision and pattern recognition},
  pages={3343--3352},
  year={2019}
}

@inproceedings{lin2022sar,
  title={{SAR-Net}: Shape alignment and recovery network for category-level {6D} object pose and size estimation},
  author={Lin, Haitao and Liu, Zichang and Cheang, Chilam and Fu, Yanwei and Guo, Guodong and Xue, Xiangyang},
  booktitle={Proceedings of the IEEE/CVF conference on computer vision and pattern recognition},
  pages={6707--6717},
  year={2022}
}

@inproceedings{zhou2019continuity,
  title={On the continuity of rotation representations in neural networks},
  author={Zhou, Yi and Barnes, Connelly and Lu, Jingwan and Yang, Jimei and Li, Hao},
  booktitle={Proceedings of the IEEE/CVF conference on computer vision and pattern recognition},
  pages={5745--5753},
  year={2019}
}

@article{zhang2023genpose,
  title={{GenPose}: Generative category-level object pose estimation via diffusion models},
  author={Zhang, Jiyao and Wu, Mingdong and Dong, Hao},
  journal={arXiv preprint arXiv:2306.10531},
  year={2023}
}

@inproceedings{wang2019normalized,
  title={Normalized object coordinate space for category-level {6D} object pose and size estimation},
  author={Wang, He and Sridhar, Srinath and Huang, Jingwei and Valentin, Julien and Song, Shuran and Guibas, Leonidas J},
  booktitle={Proceedings of the IEEE/CVF Conference on Computer Vision and Pattern Recognition},
  pages={2642--2651},
  year={2019}
}

@inproceedings{lin2023pourit,
  title={{PourIt!}: Weakly-supervised Liquid Perception from a Single Image for Visual Closed-Loop Robotic Pouring},
  author={Lin, Haitao and Fu, Yanwei and Xue, Xiangyang},
  booktitle={Proceedings of the IEEE/CVF International Conference on Computer Vision},
  pages={241--251},
  year={2023}
}

@article{wang2024polaris,
  title={{Polaris}: Open-ended Interactive Robotic Manipulation via Syn2Real Visual Grounding and Large Language Models},
  author={Wang, Tianyu and Lin, Haitao and Yu, Junqiu and Fu, Yanwei},
  journal={arXiv preprint arXiv:2408.07975},
  year={2024}
}

@inproceedings{zhang2025omni6dpose,
  title={{Omni6DPose}: A benchmark and model for universal {6D} object pose estimation and tracking},
  author={Zhang, Jiyao and Huang, Weiyao and Peng, Bo and Wu, Mingdong and Hu, Fei and Chen, Zijian and Zhao, Bo and Dong, Hao},
  booktitle={European Conference on Computer Vision},
  pages={199--216},
  year={2025},
  organization={Springer}
}

@inproceedings{ouyang2025rfmpose,
  title={{RFMPose: Generative Category-Level Object Pose Estimation via Riemannian Flow Matching}},
  author={Ouyang, Wenzhe and Ye, Qi and Wang, Jinghua and Xu, Zenglin and Chen, Jiming},
  booktitle={Advances in Neural Information Processing Systems},
  year={2025},
  url={https://mlanthology.org/neurips/2025/ouyang2025neurips-rfmpose/}
}

@article{ravi2024sam,
  title={{SAM 2}: Segment anything in images and videos},
  author={Ravi, Nikhila and Gabeur, Valentin and Hu, Yuan-Ting and Hu, Ronghang and Ryali, Chaitanya and Ma, Tengyu and Khedr, Haitham and R{\"a}dle, Roman and Rolland, Chloe and Gustafson, Laura and others},
  journal={arXiv preprint arXiv:2408.00714},
  year={2024}
}

@article{oquab2023dinov2,
  title={{DINOv2}: Learning robust visual features without supervision},
  author={Oquab, Maxime and Darcet, Timoth{\'e}e and Moutakanni, Th{\'e}o and Vo, Huy and Szafraniec, Marc and Khalidov, Vasil and Fernandez, Pierre and Haziza, Daniel and Massa, Francisco and El-Nouby, Alaaeldin and others},
  journal={arXiv preprint arXiv:2304.07193},
  year={2023}
}

@article{zhang2024tale,
  title={A tale of two features: Stable diffusion complements {DINO} for zero-shot semantic correspondence},
  author={Zhang, Junyi and Herrmann, Charles and Hur, Junhwa and Polania Cabrera, Luisa and Jampani, Varun and Sun, Deqing and Yang, Ming-Hsuan},
  journal={Advances in Neural Information Processing Systems},
  volume={36},
  year={2023}
}

@inproceedings{yu2021pointr,
  title={{PoinTr}: Diverse Point Cloud Completion with Geometry-Aware Transformers},
  author={Yu, Xumin and Rao, Yongming and Wang, Ziyi and Liu, Zuyan and Lu, Jiwen and Zhou, Jie},
  booktitle={ICCV},
  year={2021}
}

@article{fu2022category,
  title={Category-level {6D} object pose estimation in the wild: A semi-supervised learning approach and a new dataset},
  author={Fu, Yang and Wang, Xiaolong},
  journal={Advances in Neural Information Processing Systems},
  volume={35},
  pages={27469--27483},
  year={2022}
}

@inproceedings{wang2022phocal,
  title={{PhoCaL}: A multi-modal dataset for category-level object pose estimation with photometrically challenging objects},
  author={Wang, Pengyuan and Jung, HyunJun and Li, Yitong and Shen, Siyuan and Srikanth, Rahul Parthasarathy and Garattoni, Lorenzo and Meier, Sven and Navab, Nassir and Busam, Benjamin},
  booktitle={Proceedings of the IEEE/CVF conference on computer vision and pattern recognition},
  pages={21222--21231},
  year={2022}
}

@article{lin2026universal,
  title={Universal Pose Pretraining for Generalizable Vision-Language-Action Policies},
  author={Lin, Haitao and Yu, Hanyang and Huang, Jingshun and Zhang, He and Ling, Yonggen and Tan, Ping and Xue, Xiangyang and Fu, Yanwei},
  journal={arXiv preprint arXiv:2602.19710},
  year={2026}
}

@inproceedings{zhang2025omni6d,
  title={{Omni6D}: Large-Vocabulary {3D} Object Dataset for Category-Level {6D} Object Pose Estimation},
  author={Zhang, Mengchen and Wu, Tong and Wang, Tai and Wang, Tengfei and Liu, Ziwei and Lin, Dahua},
  booktitle={European Conference on Computer Vision},
  pages={216--232},
  year={2025},
  organization={Springer}
}

@inproceedings{tian2020shape,
  title={Shape prior deformation for categorical {6D} object pose and size estimation},
  author={Tian, Meng and Ang, Jr., Marcelo H. and Lee, Gim Hee},
  booktitle={Computer Vision--ECCV 2020: 16th European Conference, Glasgow, UK, August 23--28, 2020, Proceedings, Part XXI 16},
  pages={530--546},
  year={2020},
  organization={Springer}
}

@inproceedings{zheng2023hs,
  title={{HS-Pose}: Hybrid scope feature extraction for category-level object pose estimation},
  author={Zheng, Linfang and Wang, Chen and Sun, Yinghan and Dasgupta, Esha and Chen, Hua and Leonardis, Ale{\v{s}} and Zhang, Wei and Chang, Hyung Jin},
  booktitle={Proceedings of the IEEE/CVF Conference on Computer Vision and Pattern Recognition},
  pages={17163--17173},
  year={2023}
}

@inproceedings{liu2023net,
  title={{IST-Net}: Prior-free category-level pose estimation with implicit space transformation},
  author={Liu, Jianhui and Chen, Yukang and Ye, Xiaoqing and Qi, Xiaojuan},
  booktitle={Proceedings of the IEEE/CVF International Conference on Computer Vision},
  pages={13978--13988},
  year={2023}
}

@inproceedings{chen2021fs,
  title={{FS-Net}: Fast shape-based network for category-level {6D} object pose estimation with decoupled rotation mechanism},
  author={Chen, Wei and Jia, Xi and Chang, Hyung Jin and Duan, Jinming and Shen, Linlin and Leonardis, Ales},
  booktitle={Proceedings of the IEEE/CVF Conference on Computer Vision and Pattern Recognition},
  pages={1581--1590},
  year={2021}
}

@article{labbe2022megapose,
  title={{MegaPose}: {6D} pose estimation of novel objects via render \& compare},
  author={Labb{\'e}, Yann and Manuelli, Lucas and Mousavian, Arsalan and Tyree, Stephen and Birchfield, Stan and Tremblay, Jonathan and Carpentier, Justin and Aubry, Mathieu and Fox, Dieter and Sivic, Josef},
  journal={arXiv preprint arXiv:2212.06870},
  year={2022}
}

@inproceedings{wen2024foundationpose,
  title={{FoundationPose}: Unified {6D} pose estimation and tracking of novel objects},
  author={Wen, Bowen and Yang, Wei and Kautz, Jan and Birchfield, Stan},
  booktitle={Proceedings of the IEEE/CVF Conference on Computer Vision and Pattern Recognition},
  pages={17868--17879},
  year={2024}
}

@inproceedings{liu2025one2any,
  title={{One2Any}: One-Reference {6D} Pose Estimation for Any Object},
  author={Liu, Mengya and Li, Siyuan and Chhatkuli, Ajad and Truong, Prune and Van Gool, Luc and Tombari, Federico},
  booktitle={Proceedings of the Computer Vision and Pattern Recognition Conference},
  pages={6457--6467},
  year={2025}
}

@inproceedings{chen2021sgpa,
  title={{SGPA}: Structure-guided prior adaptation for category-level {6D} object pose estimation},
  author={Chen, Kai and Dou, Qi},
  booktitle={Proceedings of the IEEE/CVF International Conference on Computer Vision},
  pages={2773--2782},
  year={2021}
}

@misc{yang2018foldingnetpointcloudautoencoder,
      title={{FoldingNet}: Point Cloud Auto-encoder via Deep Grid Deformation},
      author={Yaoqing Yang and Chen Feng and Yiru Shen and Dong Tian},
      year={2018},
      eprint={1712.07262},
      archivePrefix={arXiv},
      primaryClass={cs.CV},
      url={https://arxiv.org/abs/1712.07262}, 
}

@inproceedings{wen2023bundlesdf,
  title={{BundleSDF}: Neural {6-DoF} tracking and {3D} reconstruction of unknown objects},
  author={Wen, Bowen and Tremblay, Jonathan and Blukis, Valts and Tyree, Stephen and M{\"u}ller, Thomas and Evans, Alex and Fox, Dieter and Kautz, Jan and Birchfield, Stan},
  booktitle={Proceedings of the IEEE/CVF Conference on Computer Vision and Pattern Recognition},
  pages={606--617},
  year={2023}
}

@inproceedings{jung2024housecat6d,

  title={{HouseCat6D}-A Large-Scale Multi-Modal Category Level {6D} Object Perception Dataset with Household Objects in Realistic Scenarios},

  author={Jung, HyunJun and Wu, Shun-Cheng and Ruhkamp, Patrick and Zhai, Guangyao and Schieber, Hannah and Rizzoli, Giulia and Wang, Pengyuan and Zhao, Hongcheng and Garattoni, Lorenzo and Meier, Sven and Roth, Daniel and Navab, Nassir and Busam, Benjamin},

  booktitle={Proceedings of the IEEE/CVF Conference on Computer Vision and Pattern Recognition},

  pages={22498--22508},

  year={2024}

}

@article{huang2025cap,
  title={{CAP-Net}: A Unified Network for {6D} Pose and Size Estimation of Categorical Articulated Parts from a Single {RGB-D} Image},
  author={Huang, Jingshun and Lin, Haitao and Wang, Tianyu and Fu, Yanwei and Xue, Xiangyang and Zhu, Yi},
  journal={arXiv preprint arXiv:2504.11230},
  year={2025}
}

@inproceedings{nguyen2022templates,
  title={Templates for {3D} object pose estimation revisited: Generalization to new objects and robustness to occlusions},
  author={Nguyen, Van Nguyen and Hu, Yinlin and Xiao, Yang and Salzmann, Mathieu and Lepetit, Vincent},
  booktitle={Proceedings of the IEEE/CVF conference on computer vision and pattern recognition},
  pages={6771--6780},
  year={2022}
}

@inproceedings{krishnan2024omninocs,
  title={{OmniNOCS}: A unified {NOCS} dataset and model for {3D} lifting of {2D} objects},
  author={Krishnan, Akshay and Kundu, Abhijit and Maninis, Kevis-Kokitsi and Hays, James and Brown, Matthew},
  booktitle={European Conference on Computer Vision},
  pages={127--145},
  year={2024},
  organization={Springer}
}

@article{shazeer2017outrageously,
  title={Outrageously large neural networks: The sparsely-gated mixture-of-experts layer},
  author={Shazeer, Noam and Mirhoseini, Azalia and Maziarz, Krzysztof and Davis, Andy and Le, Quoc and Hinton, Geoffrey and Dean, Jeff},
  journal={arXiv preprint arXiv:1701.06538},
  year={2017}
}

@inproceedings{du2022glam,
  title={{GLaM}: Efficient scaling of language models with mixture-of-experts},
  author={Du, Nan and Huang, Yanping and Dai, Andrew M and Tong, Simon and Lepikhin, Dmitry and Xu, Yuanzhong and Krikun, Maxim and Zhou, Yanqi and Yu, Adams Wei and Firat, Orhan and others},
  booktitle={International conference on machine learning},
  pages={5547--5569},
  year={2022},
  organization={PMLR}
}

@article{fedus2022switch,
  title={Switch transformers: Scaling to trillion parameter models with simple and efficient sparsity},
  author={Fedus, William and Zoph, Barret and Shazeer, Noam},
  journal={Journal of Machine Learning Research},
  volume={23},
  number={120},
  pages={1--39},
  year={2022}
}

@article{liu2024deepseek,
  title={{DeepSeek-V3} technical report},
  author={{DeepSeek-AI} and Liu, Aixin and Feng, Bei and Xue, Bing and Wang, Bingxuan and Wu, Bochao and Lu, Chengda and Zhao, Chenggang and Deng, Chengqi and Zhang, Chenyu and Ruan, Chong and others},
  journal={arXiv preprint arXiv:2412.19437},
  year={2024}
}

@inproceedings{heinrich2025radiov2,
  title={{RADIOv2.5}: Improved baselines for agglomerative vision foundation models},
  author={Heinrich, Greg and Ranzinger, Mike and Yin, Hongxu and Lu, Yao and Kautz, Jan and Tao, Andrew and Catanzaro, Bryan and Molchanov, Pavlo},
  booktitle={Proc. CVPR},
  pages={22487--22497},
  year={2025}
}

@article{zhang2024lapose,
  title={{LaPose}: Laplacian Mixture Shape Modeling for {RGB}-Based Category-Level Object Pose Estimation},
  author={Zhang, Ruida and Huang, Ziqin and Wang, Gu and Zhang, Chenyangguang and Di, Yan and Zuo, Xingxing and Tang, Jiwen and Ji, Xiangyang},
  journal={arXiv preprint arXiv:2409.15727},
  year={2024}
}

@inproceedings{radford2021learning,
  title={Learning transferable visual models from natural language supervision},
  author={Radford, Alec and Kim, Jong Wook and Hallacy, Chris and Ramesh, Aditya and Goh, Gabriel and Agarwal, Sandhini and Sastry, Girish and Askell, Amanda and Mishkin, Pamela and Clark, Jack and others},
  booktitle={International conference on machine learning},
  pages={8748--8763},
  year={2021},
  organization={PmLR}
}

@misc{yu2023adapointrdiversepointcloud,
      title={{AdaPoinTr}: Diverse Point Cloud Completion with Adaptive Geometry-Aware Transformers},
      author={Xumin Yu and Yongming Rao and Ziyi Wang and Jiwen Lu and Jie Zhou},
      year={2023},
      eprint={2301.04545},
      archivePrefix={arXiv},
      primaryClass={cs.CV},
      url={https://arxiv.org/abs/2301.04545}, 
}

@article{fang2023data,
  title={Data filtering networks},
  author={Fang, Alex and Jose, Albin Madappally and Jain, Amit and Schmidt, Ludwig and Toshev, Alexander and Shankar, Vaishaal},
  journal={arXiv preprint arXiv:2309.17425},
  year={2023}
}

@article{vaswani2017attention,
  title={Attention is all you need},
  author={Vaswani, Ashish and Shazeer, Noam and Parmar, Niki and Uszkoreit, Jakob and Jones, Llion and Gomez, Aidan N and Kaiser, {\L}ukasz and Polosukhin, Illia},
  journal={Advances in neural information processing systems},
  volume={30},
  year={2017}
}

@inproceedings{liu2022gen6d,
  title={{Gen6D}: Generalizable model-free {6-DoF} object pose estimation from {RGB} images},
  author={Liu, Yuan and Wen, Yilin and Peng, Sida and Lin, Cheng and Long, Xiaoxiao and Komura, Taku and Wang, Wenping},
  booktitle={European Conference on Computer Vision},
  pages={298--315},
  year={2022},
  organization={Springer}
}

@article{he2022onepose++,
  title={{OnePose++}: Keypoint-free one-shot object pose estimation without {CAD} models},
  author={He, Xingyi and Sun, Jiaming and Wang, Yuang and Huang, Di and Bao, Hujun and Zhou, Xiaowei},
  journal={Advances in Neural Information Processing Systems},
  volume={35},
  pages={35103--35115},
  year={2022}
}

@inproceedings{he2022fs6d,
  title={{FS6D}: Few-shot {6D} pose estimation of novel objects},
  author={He, Yisheng and Wang, Yao and Fan, Haoqiang and Sun, Jian and Chen, Qifeng},
  booktitle={Proceedings of the IEEE/CVF Conference on Computer Vision and Pattern Recognition},
  pages={6814--6824},
  year={2022}
}

@inproceedings{li2023nerf,
  title={{NeRF-Pose}: A first-reconstruct-then-regress approach for weakly-supervised {6D} object pose estimation},
  author={Li, Fu and Vutukur, Shishir Reddy and Yu, Hao and Shugurov, Ivan and Busam, Benjamin and Yang, Shaowu and Ilic, Slobodan},
  booktitle={Proceedings of the IEEE/CVF International Conference on Computer Vision},
  pages={2123--2133},
  year={2023}
}

@inproceedings{nguyen2024gigapose,
  title={{GigaPose}: Fast and robust novel object pose estimation via one correspondence},
  author={Nguyen, Van Nguyen and Groueix, Thibault and Salzmann, Mathieu and Lepetit, Vincent},
  booktitle={Proceedings of the IEEE/CVF Conference on Computer Vision and Pattern Recognition},
  pages={9903--9913},
  year={2024}
}

@inproceedings{irshad2022shapo,
  title={{ShAPO}: Implicit representations for multi-object shape, appearance, and pose optimization},
  author={Irshad, Muhammad Zubair and Zakharov, Sergey and Ambru{\textcommabelow{s}}, Rare{\textcommabelow{s}} and Kollar, Thomas and Kira, Zsolt and Gaidon, Adrien},
  booktitle={European Conference on Computer Vision},
  pages={275--292},
  year={2022},
  organization={Springer}
}

@article{huang2025you,
  title={You Only Estimate Once: Unified, One-stage, Real-Time Category-level Articulated Object {6D} Pose Estimation for Robotic Grasping},
  author={Huang, Jingshun and Lin, Haitao and Wang, Tianyu and Fu, Yanwei and Jiang, Yu-Gang and Xue, Xiangyang},
  journal={arXiv preprint arXiv:2506.05719},
  year={2025}
}

@inproceedings{guo2023handal,
  title={{HANDAL}: A dataset of real-world manipulable object categories with pose annotations, affordances, and reconstructions},
  author={Guo, Andrew and Wen, Bowen and Yuan, Jianhe and Tremblay, Jonathan and Tyree, Stephen and Smith, Jeffrey and Birchfield, Stan},
  booktitle={2023 IEEE/RSJ International Conference on Intelligent Robots and Systems (IROS)},
  pages={11428--11435},
  year={2023},
  organization={IEEE}
}

@inproceedings{lee2025any6d,
    title     = {{Any6D}: Model-free {6D} Pose Estimation of Novel Objects},
    author    = {Lee, Taeyeop and Wen, Bowen and Kang, Minjun and Kang, Gyuree and Kweon, In So and Yoon, Kuk-Jin},
    booktitle = {Proceedings of the Computer Vision and Pattern Recognition Conference (CVPR)},
    year      = {2025},
}

@misc{lee2025jointlearningposeregression,
      title={Joint Learning of Pose Regression and Denoising Diffusion with Score Scaling Sampling for Category-level {6D} Pose Estimation},
      author={Seunghyun Lee and Tae-Kyun Kim},
      year={2025},
      eprint={2510.04125},
      archivePrefix={arXiv},
      primaryClass={cs.CV},
      url={https://arxiv.org/abs/2510.04125}, 
}

@misc{sam3d2025,
      title={{SAM 3D}: 3Dfy Anything in Images},
      author={{SAM 3D Team} and Xingyu Chen and Fu-Jen Chu and Pierre Gleize and Kevin J. Liang and Alexander Sax and Hao Tang and Weiyao Wang and Michelle Guo and Thibaut Hardin and Xiang Li and Aohan Lin and Jiawei Liu and Ziqi Ma and Anushka Sagar and Bowen Song and Xiaodong Wang and Jianing Yang and Bowen Zhang and Piotr Doll{\'a}r and Georgia Gkioxari and Matt Feiszli and Jitendra Malik},
      year={2025},
      eprint={2511.16624},
      archivePrefix={arXiv},
      primaryClass={cs.CV},
      url={https://arxiv.org/abs/2511.16624},
}

@article{tang20193d,
  title={{3D} mapping and {6D} pose computation for real time augmented reality on cylindrical objects},
  author={Tang, Fulin and Wu, Yihong and Hou, Xiaohui and Ling, Haibin},
  journal={IEEE Transactions on Circuits and Systems for Video Technology},
  volume={30},
  number={9},
  pages={2887--2899},
  year={2019},
  publisher={IEEE}
}

@inproceedings{durchon20246d,
  title={{6D} pose estimation of unseen objects for industrial augmented reality},
  author={Durchon, Hugo and Preda, Marius and Zaharia, Titus},
  booktitle={2024 IEEE 20th International Conference on Intelligent Computer Communication and Processing (ICCP)},
  pages={1--8},
  year={2024},
  organization={IEEE}
}

@article{tan20176d,
  title={{6D} object pose estimation with depth images: A seamless approach for robotic interaction and augmented reality},
  author={Tan, David Joseph and Navab, Nassir and Tombari, Federico},
  journal={arXiv preprint arXiv:1709.01459},
  year={2017}
}

@inproceedings{chen2024secondpose,
  author={Chen, Yamei and Di, Yan and Zhai, Guangyao and Manhardt, Fabian and Zhang, Chenyangguang and Zhang, Ruida and Tombari, Federico and Navab, Nassir and Busam, Benjamin},
  title={{SecondPose: SE(3)-Consistent Dual-Stream Feature Fusion for Category-Level Pose Estimation}},
  booktitle={Proceedings of the IEEE/CVF Conference on Computer Vision and Pattern Recognition},
  year={2024},
  pages={9959--9969}
}


\appendix

\section{Additional Implementation Details}

\subsection{Model Details}
\label{app:implement}

F2G-Pose receives a segmented RGB crop and its corresponding partial point cloud. We sample 1024 points from the segmented depth observation and concatenate each point coordinate with a lifted visual feature. Frozen RADIOv2.5-L~\cite{heinrich2025radiov2} provides feature maps from layers 8, 16, and 23. Independent $1\times1$ convolutional blocks compress each map from 1024 to 512 channels. Learned softmax-normalized layer weights combine them by summation, followed by a $3\times3$ convolutional block producing 1024 channels; both convolutional stages use BatchNorm and ReLU. We gather each point's feature by integer-dividing its crop-pixel coordinates by the patch size of 16. This produces an input tensor of size \(B \times 1024 \times 1027\).

The fused RGB-D point set is tokenized by a DGCNN encoder with two downsampling stages, reducing 1024 input points to 128 object tokens. The tokens are projected to 512 dimensions and processed by 8 Transformer blocks with 8 attention heads. In the full model, each Transformer feed-forward layer is replaced by a top-2 Mixture-of-Experts layer with 8 experts and the load-balancing loss described in the main paper. A learned softmax-weighted sum combines the block outputs; an MLP projects the tokens to 1024 dimensions before max pooling. The pose-size branch directly regresses rotation, translation, and metric size. The shape branch predicts 512 initial points, concatenates 256 farthest-point-sampled observed points, scores the 768 candidates and retains 512, then predicts four local offsets per retained point for a 2048-point completed shape. The selected 512 points and final 2048 points receive reconstruction supervision. Continuous rotational symmetries use $1^\circ$ candidate sampling; objects with at least two continuously symmetric axes are excluded from the rotation loss.

\subsection{Training Details}
\label{app:train}

F2G-Pose is trained only on the SOPE training split used in the main experiments. We train for 100 epochs with batch size 512 using AdamW. The initial learning rate is \(10^{-4}\), the weight decay is \(5 \times 10^{-4}\), and a LambdaLR schedule decays the learning rate by a factor of 0.9 every 8 epochs with a lower bound of 2\% of the initial value. BatchNorm momentum is decayed from 0.9 by a factor of 0.5 every 3 epochs with a minimum value of 0.01. Training experiments are run with PyTorch 2.4 and CUDA 12.4 using automatic mixed precision, DistributedDataParallel, and synchronized BatchNorm on 4 NVIDIA H200 GPUs.

For runtime comparison, we benchmark executable methods under a unified inference protocol using the same segmentation masks, RGB-D crops, depth preprocessing, batch size 1, and GPU-synchronized timing on one RTX 4090. The pose-only setting disables the shape head after the shared RGB-D encoder, while the full setting keeps the shape branch enabled.

\subsection{Controlled Robustness Evaluation}
\label{app:controlled_robustness}

To complement the material robustness study in the main paper, we evaluate controlled perturbations on ROPE. RGB perturbations simulate common photometric changes, while depth perturbations simulate sensor noise. These results are intended as stress tests of the current SOPE-trained model rather than as a separate dataset contribution.

\paragraph{Photometric perturbations.}
For RGB inputs, we apply exposure shifts (\(\pm 0.5 / \pm 1.3 / \pm 2.2\) EV), white-balance shifts (\(\pm 800 / \pm 2000 / \pm 4000\) K), and synthetic illumination changes (gradient or vignetting at \(0.10 / 0.25 / 0.40\) strength). Table~\ref{tab:photometric_robustness} shows that performance changes only mildly under light and moderate perturbations, while heavy perturbations produce a larger but still bounded drop.

\paragraph{Depth perturbations.}
For depth inputs, we inject heteroscedastic noise
\[
\sigma(z) = \alpha + \beta z^2,
\]
where \(\alpha \in \{0.001, 0.002, 0.003\}\) and \(\beta \in \{0.000, 0.003, 0.006\}\). As summarized in Table~\ref{tab:depth_robustness}, depth noise affects the method more strongly than photometric shifts, consistent with the main paper's observation that precise metric alignment is sensitive to corrupted or missing depth.

\begin{figure}[htb]
    \centering
    \includegraphics[width=0.35\textwidth]{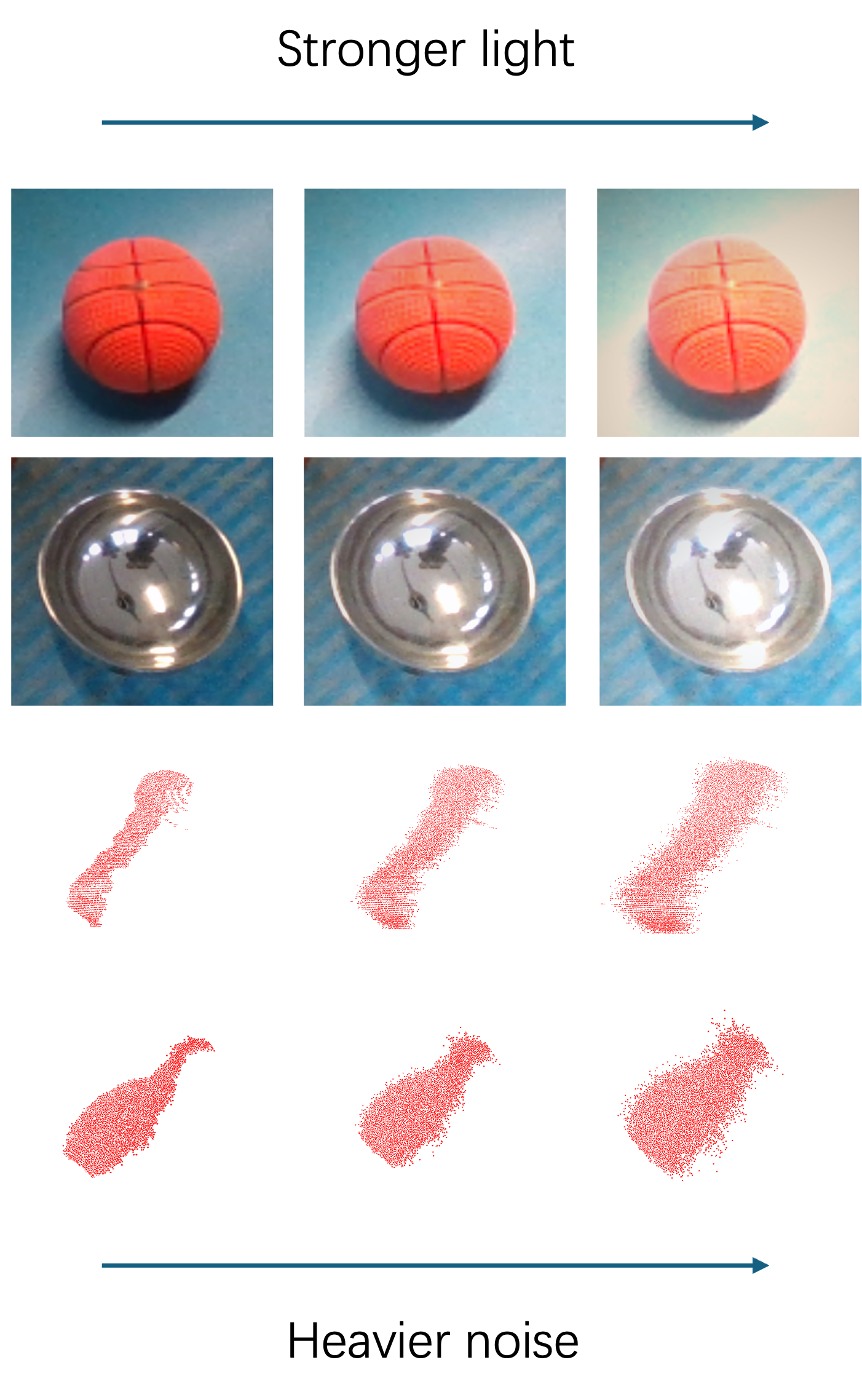}
    \caption{Examples of controlled RGB and depth perturbations.}
    \label{fig:perturbation}
\end{figure}

\begin{table}[htb]
    \centering
    \caption{Controlled robustness under photometric perturbations on ROPE.}
    \label{tab:photometric_robustness}
    \resizebox{\textwidth}{!}{
    \begin{tabular}{lcccccccc}
    \toprule
    Perturbation & IoU25 & IoU50 & IoU75 & 5$^\circ$2cm & 5$^\circ$5cm & 10$^\circ$2cm & 10$^\circ$5cm & $\Delta$AUC / $\Delta$VUS (\%) \\
    \midrule
    Clean & 46.5 & 27.8 & 5.8 & 11.3 & 15.9 & 21.8 & 31.1 & +0.0 / +0.0 \\
    Light & 46.3 & 27.7 & 5.7 & 11.3 & 15.9 & 21.7 & 30.8 & -0.5 / -0.6 \\
    Moderate & 46.1 & 27.4 & 5.6 & 11.2 & 15.7 & 21.5 & 30.5 & -1.3 / -1.6 \\
    Heavy & 44.7 & 26.4 & 5.4 & 10.6 & 15.2 & 20.5 & 29.1 & -4.7 / -6.4 \\
    \bottomrule
    \end{tabular}}
\end{table}

\begin{table}[htb]
    \centering
    \caption{Controlled robustness under heteroscedastic depth noise on ROPE.}
    \label{tab:depth_robustness}
    \resizebox{\textwidth}{!}{
    \begin{tabular}{lcccccccccc}
    \toprule
    Perturbation & $\alpha$ & $\beta$ & IoU25 & IoU50 & IoU75 & 5$^\circ$2cm & 5$^\circ$5cm & 10$^\circ$2cm & 10$^\circ$5cm & $\Delta$AUC / $\Delta$VUS (\%) \\
    \midrule
    Clean & 0.000 & 0.000 & 46.5 & 27.8 & 5.8 & 11.3 & 15.9 & 21.8 & 31.1 & +0.0 / +0.0 \\
    Low & 0.001 & 0.000 & 46.3 & 27.7 & 5.7 & 11.2 & 15.9 & 21.9 & 31.2 & -0.5 / +0.1 \\
    Moderate & 0.002 & 0.003 & 43.3 & 24.4 & 4.5 & 10.1 & 14.9 & 20.3 & 29.5 & -10.4 / -7.2 \\
    Heavy & 0.003 & 0.006 & 39.0 & 20.0 & 3.3 & 8.5 & 13.2 & 17.6 & 26.9 & -23.4 / -20.0 \\
    \bottomrule
    \end{tabular}}
\end{table}

\subsection{Additional GenPose++ Candidate Diagnostics}
\label{app:candidate_diagnostics}

Table~\ref{tab:genpose_symmetry_app} breaks down the GenPose++ candidate set by object symmetry. The oracle best-of-50 result in the main paper indicates that the candidate set often contains accurate hypotheses, but the breakdown shows that candidate quality is uneven across object structures. Objects without rotational symmetry are especially difficult: they have fewer near-perfect candidates and a larger average rotation error, so post-hoc scoring or clustering has a harder selection problem.

\begin{table}[htb]
\centering
\caption{GenPose++ candidate quality by symmetry on ROPE. Values are computed over sampled candidates before the final candidate extraction step.}
\label{tab:genpose_symmetry_app}
\small
\setlength\tabcolsep{5pt}
\renewcommand{\arraystretch}{1.0}
\begin{tabular}{lcc}
\toprule
Symmetry type & Mean rotation error $\downarrow$ & Candidates $\leq 5^\circ$ $\uparrow$ \\
\midrule
Continuous & 24.0 & 49\% \\
Quarter-turn & 19.0 & 33\% \\
Half-turn & 45.0 & 25\% \\
None & 57.0 & 21\% \\
\bottomrule
\end{tabular}
\end{table}

\subsection{Failure Mode Analysis}
\label{app:failure_modes}

The qualitative failure cases are drawn from ROPE and are consistent with the limitations stated in the paper: corrupted depth on thin or reflective structures, heavy occlusion, and ambiguous rotations for symmetric or weakly textured objects.

\begin{figure}[htb]
    \centering
    \includegraphics[width=1.0\textwidth]{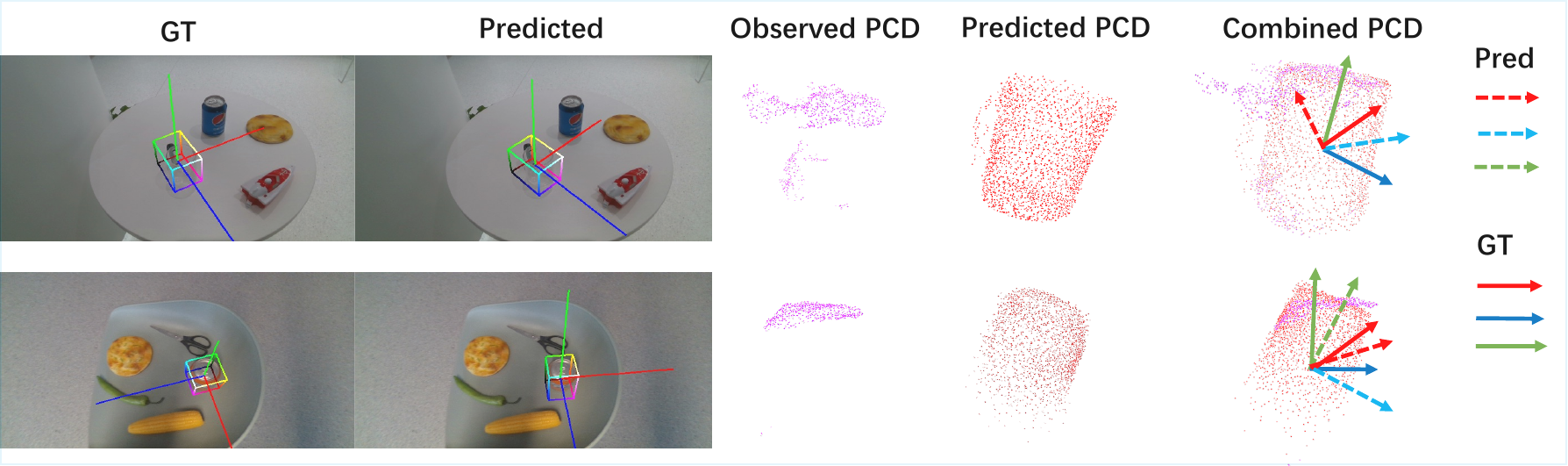}
    \caption{Failure cases caused by corrupted or incomplete depth on reflective object regions.}
    \label{fig:failure}
\end{figure}

\paragraph{Reflective and transparent materials.}
Reflective or transparent surfaces often produce missing or distorted depth. RADIO features provide complementary appearance cues, but they cannot fully recover metric extent when the observed point cloud loses thin handles, sharp edges, or asymmetric parts. This leads to high-precision IoU and rotation errors, as shown in Fig.~\ref{fig:failure}.

\paragraph{Occlusion.}
Under mild or moderate occlusion, the model can often infer pose from visible geometry and lifted semantic cues. Severe occlusion removes too much task-relevant evidence, especially when the hidden region contains the asymmetric part needed to disambiguate orientation.

\begin{figure}[htb]
    \centering
    \includegraphics[width=0.85\textwidth]{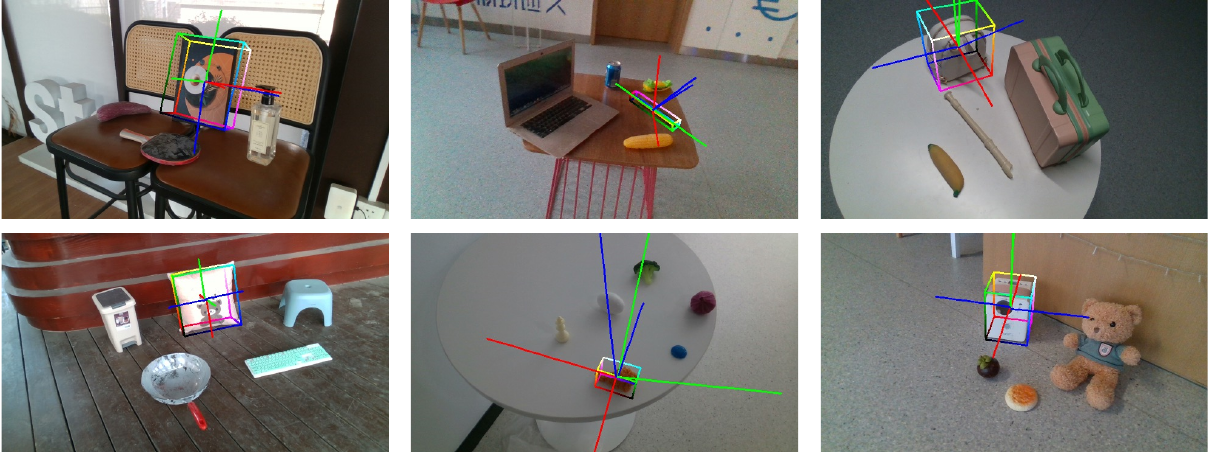}
    \caption{Failure cases caused by symmetric, geometrically regular, or small objects with ambiguous orientation cues.}
    \label{fig:ambiguity}
\end{figure}

\paragraph{Category ambiguity.}
For symmetric, regular, or small low-texture objects, multiple rotations can be visually plausible from a single RGB-D observation. The model may therefore predict a symmetry-related or near-symmetry-related orientation, producing typical \(90^\circ\) or \(180^\circ\) errors as illustrated in Fig.~\ref{fig:ambiguity}.

\subsection{Supplementary Ablation Results}
\label{app:ablation}

\begin{table}[htb]
\caption{Supplementary pose-size ablations on ROPE. We keep the metric definitions consistent with the main tables; ``--'' indicates a metric that is not applicable or was not measured for that variant.}
\label{tab:ablation_summary}
\centering
\footnotesize
\setlength\tabcolsep{2pt}
\renewcommand{\arraystretch}{1.0}
\resizebox{\linewidth}{!}{
\begin{tabular}{lccccc}
\toprule
\textbf{Setting} & \textbf{AUC IoU25} & \textbf{AUC IoU50} & \textbf{AUC IoU75} & \textbf{VUS 5$^\circ$5cm} & \textbf{VUS 10$^\circ$2cm} \\
\midrule
\multicolumn{6}{c}{\textit{Main components}} \\
\midrule
Full & 46.5 & 27.8 & 5.8 & 15.9 & 21.8 \\
w/o RGB & 34.5 & 16.2 & 2.2 & 6.5 & 13.2 \\
w/o MoE & 41.0 & 24.0 & 3.9 & 8.7 & 17.8 \\
w/o Shape Sup. & 39.5 & 23.2 & 3.5 & 8.1 & 16.5 \\
\midrule
\multicolumn{6}{c}{\textit{MoE routing and capacity}} \\
\midrule
MoE top-1 & -- & 25.6 & -- & 13.5 & -- \\
MoE top-2, no balance & -- & 26.0 & -- & 14.2 & -- \\
MoE top-4 & -- & 28.0 & -- & 16.1 & -- \\
\midrule
\multicolumn{6}{c}{\textit{Backbone and shape-decoder variants}} \\
\midrule
Full + DINOv2 & 45.5 & 27.0 & -- & 13.7 & -- \\
Full + CLIP & 44.8 & 26.3 & -- & 12.5 & -- \\
w/o selection & 45.7 & 27.1 & 5.2 & -- & 21.0 \\
w/o coarse-to-fine & 43.8 & 25.6 & 4.3 & -- & 19.9 \\
\bottomrule
\end{tabular}
}
\end{table}

Table~\ref{tab:ablation_summary} summarizes additional ablations while avoiding new claims beyond the current method. The RGB ablation confirms that lifted foundation features are a major contributor under partial depth. The MoE variants show that top-2 routing is the adopted trade-off: top-4 gives only a marginal gain over top-2 at higher cost, while top-1 is faster but less accurate. The backbone variants show that the framework remains effective with DINOv2 or CLIP, although RADIO gives the best result. The shape-decoder variants indicate that point selection and coarse-to-fine refinement also affect pose-size accuracy through the shared representation.

\subsection{Additional Qualitative Results}
\label{app:qualitative}

\begin{figure}[htb]
    \centering
    \includegraphics[width=0.85\textwidth]{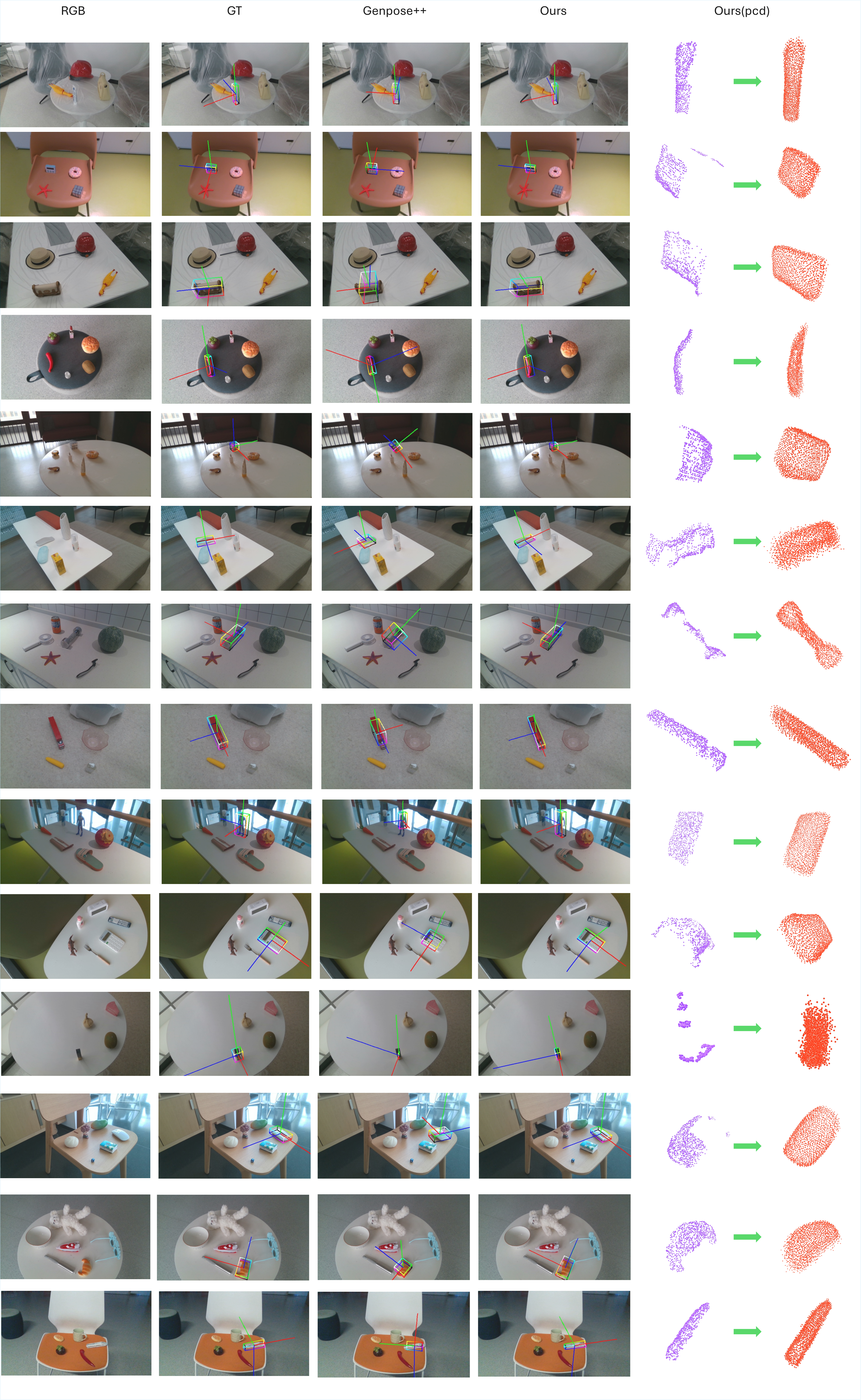}
    \caption{Additional qualitative examples from ROPE.}
    \label{fig:exemplar}
\end{figure}

Fig.~\ref{fig:exemplar} provides additional ROPE examples with the input RGB image, ground-truth annotation, GenPose++ prediction, and F2G-Pose prediction.

\begin{figure}[htb]
    \centering
    \includegraphics[width=0.8\textwidth]{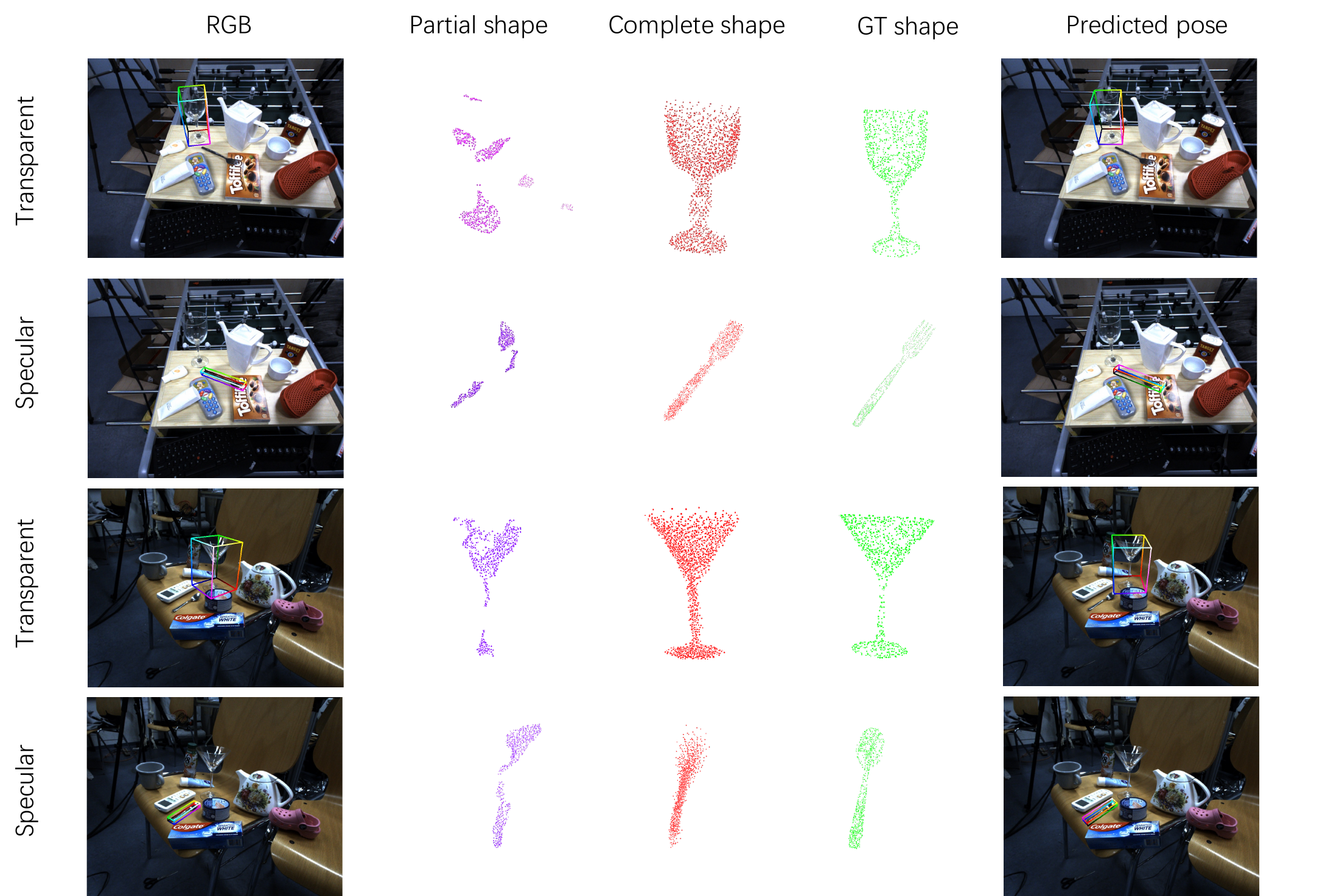}
    \caption{Reflective and transparent object examples from HouseCat6D. These cases illustrate material-induced depth corruption and the resulting camera-frame shape predictions.}
    \label{fig:special_case}
\end{figure}

\begin{figure}[htb]
    \centering
    \includegraphics[width=0.8\textwidth]{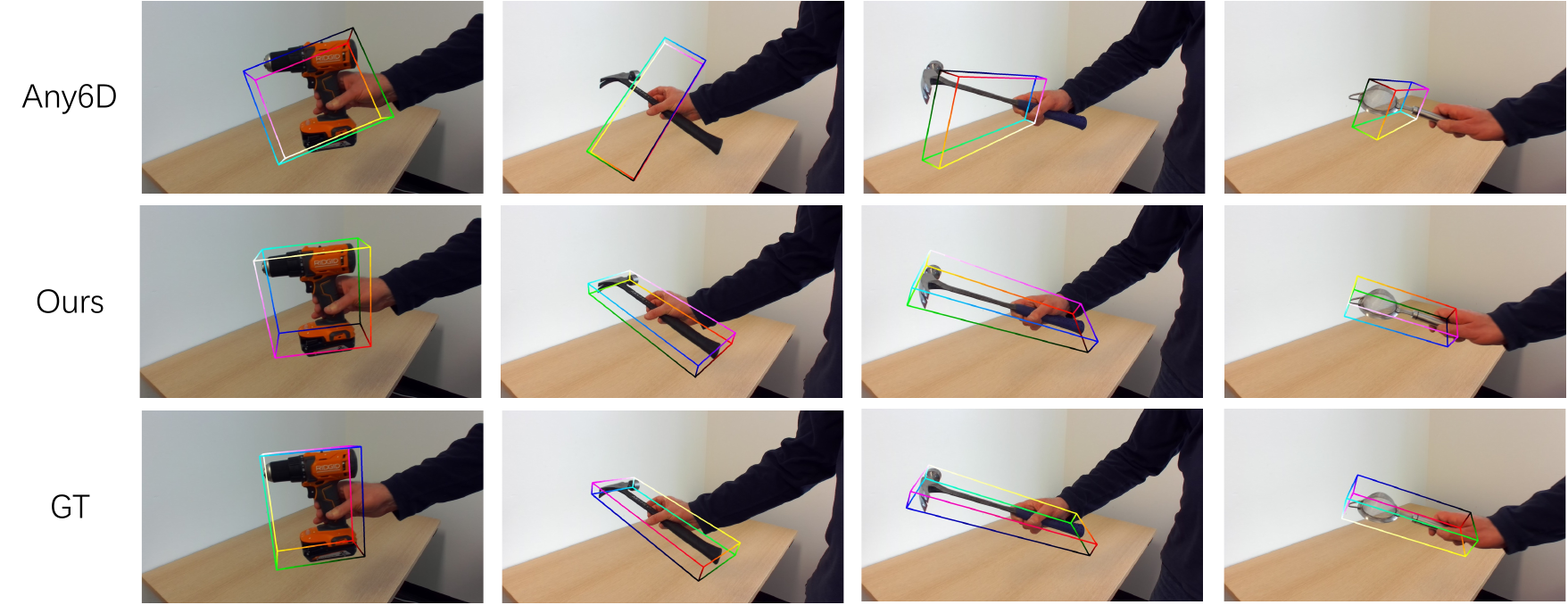}
    \caption{Qualitative results on HANDAL. We compare ground-truth 3D boxes with predictions from Any6D and F2G-Pose.}
    \label{fig:handal_cases}
\end{figure}

Fig.~\ref{fig:special_case} supplements the material robustness analysis, and Fig.~\ref{fig:handal_cases} provides additional cross-dataset real-world examples.

\clearpage
\section{Additional Evaluations and Protocol Clarifications}

\subsection{Matched RGB-D Shape Completion}
\label{app:controlled_completion}
The camera-frame comparison in Table~\ref{table:cdl1_rope_sope} measures the complete prediction system: its Chamfer distance includes intrinsic reconstruction error and global pose, translation, and scale error. Moreover, the geometry-only baselines and F2G-Pose receive different input information and supervision. To control these factors, we compare F2G-Pose and AdaPoinTr using identical RGB-D inputs, including features from the same frozen RADIO encoder, and predict shapes directly in a normalized canonical frame. Each architecture is evaluated with and without pose-size supervision; no ground-truth pose or scale is supplied as input.

\begin{table}[htbp]
\centering
\caption{Matched RGB-D completion on SOPE. All variants predict normalized canonical shapes. Canonical CD-L1 is shown $\times10^3$; rotation and translation errors are in degrees and cm. ``--'' denotes no evaluated pose output.}
\label{tab:matched_rgbd_completion}
\small
\setlength{\tabcolsep}{3pt}
\begin{tabular}{clcrrrrrr}
\toprule
& & Pose/ & & \multicolumn{2}{c}{VUS $\uparrow$} & \multicolumn{2}{c}{Mean error $\downarrow$} & Canon. \\
Group & Model & size loss & IoU50 & 5$^\circ$2cm & 10$^\circ$5cm & Rot. & Trans. & CD $\downarrow$ \\
\midrule
A & F2G-Pose & Yes & 41.5 & 18.5 & 38.5 & 17.8 & 1.10 & 34.1 \\
B & F2G-Pose & No & -- & -- & -- & -- & -- & 36.7 \\
C & AdaPoinTr & No & -- & -- & -- & -- & -- & 37.2 \\
D & AdaPoinTr & Yes & 38.5 & 16.5 & 35.0 & 19.5 & 1.25 & 35.5 \\
\bottomrule
\end{tabular}
\end{table}

Table~\ref{tab:matched_rgbd_completion} shows that pose-size supervision improves canonical completion for both architectures: $36.7\rightarrow34.1$ for F2G-Pose and $37.2\rightarrow35.5$ for AdaPoinTr. Without this supervision, F2G-Pose retains a modest numerical advantage ($36.7$ versus $37.2$). Under full supervision, it also achieves higher pose accuracy ($41.5$ versus $38.5$ IoU50 AUC). Changing our default camera-frame completion target to normalized canonical completion reduces pose IoU50 AUC from 43.7 to 41.5, supporting the camera-frame target for the pose objective. Canonical and camera-frame CD values use different coordinate systems and scales and are not directly comparable.

\subsection{Segmentation-Mask Sensitivity}
\label{app:mask_sensitivity}
We evaluate SAM-predicted masks on ROPE and controlled erosion/dilation of ground-truth masks for pose estimation on ROPE and shape completion on SOPE. Erosion removes foreground evidence, while dilation includes background. Tables~\ref{tab:mask_pose_cr} and~\ref{tab:mask_shape_cr} report the results at the stated mask-IoU levels.

\begin{table}[htbp]
\centering
\caption{Sensitivity of pose estimation to segmentation masks on ROPE. Rotation and translation columns report mean errors.}
\label{tab:mask_pose_cr}
\small
\setlength{\tabcolsep}{3pt}
\begin{tabular}{lrrrrr}
\toprule
Mask & IoU50 AUC & IoU75 AUC & VUS 5$^\circ$5cm & Rot. ($^\circ$) & Trans. (cm) \\
\midrule
GT & 27.8 & 5.8 & 15.9 & 27.5 & 1.19 \\
SAM & 27.2 & 5.6 & 15.2 & 28.1 & 1.21 \\
Erosion, IoU 90\% & 26.6 & 5.3 & 14.8 & 28.3 & 1.24 \\
Erosion, IoU 80\% & 23.8 & 4.1 & 13.7 & 28.8 & 1.32 \\
Dilation, IoU 90\% & 27.5 & 5.3 & 15.8 & 27.6 & 1.20 \\
Dilation, IoU 80\% & 25.8 & 4.0 & 15.5 & 27.8 & 1.21 \\
\bottomrule
\end{tabular}
\end{table}

\begin{table}[htbp]
\centering
\caption{Sensitivity of SOPE camera-frame shape completion to segmentation masks. Symmetric CD-L1 is shown $\times10^3$.}
\label{tab:mask_shape_cr}
\small
\begin{tabular}{lrr}
\toprule
Mask & Mask IoU & Shape CD $\downarrow$ \\
\midrule
GT & 100\% & 5.42 \\
Dilation & 90\% & 5.45 \\
Dilation & 80\% & 5.48 \\
Erosion & 90\% & 5.57 \\
Erosion & 80\% & 5.80 \\
\bottomrule
\end{tabular}
\end{table}

SAM masks yield results close to the GT-mask baseline. At matched mask IoU, erosion causes greater degradation than dilation on most pose metrics and on shape CD, suggesting that missing foreground is generally more harmful than limited background inclusion. Together with the RGB and depth perturbations in Appendix~\ref{app:controlled_robustness}, these results characterize input sensitivity; they do not remove the requirement for segmented RGB-D input.

\subsection{MoE Expert Preferences}
\label{app:moe_specialization}
Beyond load balance, we analyze instance-level routing over 7,466 object samples. Routing exhibits an association with category (Cram\'er's $V=0.306$, $p<0.001$). Different experts show preferences for broad geometry-related groups, including bag-like, flat, and bowl-like objects, with substantial overlap. This suggests distributed specialization rather than a rigid one-expert-per-category partition. Associations with material and visibility are weaker, and no significant independent association with symmetry was found. These observations complement the capacity and balance analysis in Table~\ref{tab:moe_routing}.

\subsection{HANDAL Selection and Per-Category Results}
\label{app:handal_categories}
HANDAL is an independent real-world dataset with sensing conditions, capture environments, and object instances distinct from SOPE. Among its 17 categories, we evaluate the seven with direct semantic correspondences to the SOPE taxonomy. Selection is based on taxonomy correspondence, not model performance. The remaining categories have ambiguous mappings or no direct SOPE counterparts and are excluded because no corresponding canonical frame is defined. These seven categories map to six SOPE categories, with both wrench types mapped to \emph{spanner}.

\begin{table}[htbp]
\centering
\caption{HANDAL per-category IoU AUC for the seven directly matched categories. Counts refer to instances. The last row is the unweighted macro-average over seven HANDAL categories.}
\label{tab:handal_categories_cr}
\small
\setlength{\tabcolsep}{3pt}
\begin{tabular}{llrrrr}
\toprule
HANDAL category & SOPE category & Instances & IoU25 & IoU50 & IoU75 \\
\midrule
Claw Hammer & hammer & 19 & 35.8 & 11.5 & 0.1 \\
Adjustable Wrench & spanner & 13 & 12.1 & 1.0 & 0.0 \\
Combination Wrench & spanner & 12 & 11.8 & 0.9 & 0.0 \\
Ladle & spoon & 12 & 29.4 & 3.9 & 0.0 \\
Measuring Cup & cup & 12 & 45.2 & 17.0 & 0.3 \\
Mug & mug & 12 & 68.3 & 35.4 & 1.6 \\
Pot/Pan & pan & 12 & 42.4 & 12.3 & 0.1 \\
\midrule
Macro-average & & 92 total & 35.0 & 11.7 & 0.3 \\
\bottomrule
\end{tabular}
\end{table}

The evaluation covers 92 instances. The macro-average AUCs are 35.0, 11.7, and 0.3 at IoU25, IoU50, and IoU75. These results indicate coarse 3D alignment under cross-dataset transfer, while precise joint estimation of pose, size, and extent remains challenging. Both ROPE and this HANDAL subset evaluate seen-category, unseen-instance transfer, rather than unseen-category generalization.

\subsection{Reference-Free and Single-Reference Evaluation Protocols}
\label{app:reference_protocols}
\paragraph{Any6D (zero reference).}
``Zero reference'' means that no additional reference view is supplied beyond the test RGB-D observation. We use the official repository's reference-free image-to-3D pathway, which reconstructs a coarse shape from the test observation and subsequently estimates pose using that reconstructed shape.

\paragraph{Any6D (one reference).}
We randomly sample one additional view of the same object from its ROPE multi-view sequence as the reference image and apply the standard reference-query inference protocol.

\paragraph{FoundationPose (one reference).}
We evaluate the model-free pathway under the same one-reference constraint. One additional real view of the same object is randomly sampled from its ROPE sequence; an object mesh is reconstructed from that view and used for pose estimation.

\clearpage
\ifdefined\FtwoGArxiv\else
\section*{NeurIPS Paper Checklist}

\begin{enumerate}

\item {\bf Claims}
    \item[] Question: Do the main claims made in the abstract and introduction accurately reflect the paper's contributions and scope?
    \item[] Answer: \answerYes{}
    \item[] Justification: The abstract and introduction clearly state three contributions (direct feed-forward framework, geometry-aware feature lifting with MoE Transformer, and extensive evaluation including robotic deployment), all supported by experimental results in Sections~4--5.

\item {\bf Limitations}
    \item[] Question: Does the paper discuss the limitations of the work performed by the authors?
    \item[] Answer: \answerYes{}
    \item[] Justification: We discuss limitations in the Conclusion section, including sensitivity to segmentation mask failures, thin/asymmetric parts, severe occlusion, symmetric low-texture objects, and cross-dataset domain shift.

\item {\bf Theory assumptions and proofs}
    \item[] Question: For each theoretical result, does the paper provide the full set of assumptions and a complete (and correct) proof?
    \item[] Answer: \answerNA{}
    \item[] Justification: This paper is primarily an empirical methods paper and does not include formal theoretical results or proofs.

    \item {\bf Experimental result reproducibility}
    \item[] Question: Does the paper fully disclose all the information needed to reproduce the main experimental results of the paper to the extent that it affects the main claims and/or conclusions of the paper (regardless of whether the code and data are provided or not)?
    \item[] Answer: \answerYes{}
    \item[] Justification: We describe the model architecture, training data (SOPE), loss functions, hyperparameters, optimizer, and evaluation metrics in sufficient detail. The datasets used are publicly available benchmarks.

\item {\bf Open access to data and code}
    \item[] Question: Does the paper provide open access to the data and code, with sufficient instructions to faithfully reproduce the main experimental results, as described in supplemental material?
    \item[] Answer: \answerYes{}
    \item[] Justification: The project page linked in the abstract provides public code and pretrained models, with installation, training, and evaluation instructions. The datasets are publicly available benchmarks.

\item {\bf Experimental setting/details}
    \item[] Question: Does the paper specify all the training and test details (e.g., data splits, hyperparameters, how they were chosen, type of optimizer) necessary to understand the results?
    \item[] Answer: \answerYes{}
    \item[] Justification: The paper specifies datasets, training splits, model architecture details, loss functions, evaluation metrics, optimizer, learning rate, and training schedule in the main text and appendix.

\item {\bf Experiment statistical significance}
    \item[] Question: Does the paper report error bars suitably and correctly defined or other appropriate information about the statistical significance of the experiments?
    \item[] Answer: \answerNo{}
    \item[] Justification: We do not report error bars or confidence intervals, as the evaluation follows standard benchmark protocols with fixed train/test splits, consistent with prior work in this area.

\item {\bf Experiments compute resources}
    \item[] Question: For each experiment, does the paper provide sufficient information on the computer resources (type of compute workers, memory, time of execution) needed to reproduce the experiments?
    \item[] Answer: \answerYes{}
\item[] Justification: We report inference speed (FPS) and provide compute resource information including GPU types and software versions in the appendix.

\item {\bf Code of ethics}
    \item[] Question: Does the research conducted in the paper conform, in every respect, with the NeurIPS Code of Ethics \url{https://neurips.cc/public/EthicsGuidelines}?
    \item[] Answer: \answerYes{}
    \item[] Justification: This research conforms with the NeurIPS Code of Ethics. It involves standard computer vision benchmarks and does not raise ethical concerns.

\item {\bf Broader impacts}
    \item[] Question: Does the paper discuss both potential positive societal impacts and negative societal impacts of the work performed?
    \item[] Answer: \answerYes{}
    \item[] Justification: We discuss broader impacts in the Conclusion section, noting positive applications (robotics, augmented reality) and potential negative uses (surveillance without adequate oversight).

\item {\bf Safeguards}
    \item[] Question: Does the paper describe safeguards that have been put in place for responsible release of data or models that have a high risk for misuse (e.g., pre-trained language models, image generators, or scraped datasets)?
    \item[] Answer: \answerNA{}
    \item[] Justification: The paper does not release pre-trained generative models or scraped datasets that pose a high risk for misuse. The method is a task-specific pose estimator with limited dual-use risk.

\item {\bf Licenses for existing assets}
    \item[] Question: Are the creators or original owners of assets (e.g., code, data, models), used in the paper, properly credited and are the license and terms of use explicitly mentioned and properly respected?
    \item[] Answer: \answerYes{}
    \item[] Justification: We cite the original papers for all datasets (SOPE, ROPE, HouseCat6D, HANDAL) and models (RADIOv2.5-L, DGCNN) used in this work.

\item {\bf New assets}
    \item[] Question: Are new assets introduced in the paper well documented and is the documentation provided alongside the assets?
    \item[] Answer: \answerYes{}
    \item[] Justification: The released code and pretrained model are accompanied by setup, usage, evaluation, and model documentation. No new dataset is introduced.

\item {\bf Crowdsourcing and research with human subjects}
    \item[] Question: For crowdsourcing experiments and research with human subjects, does the paper include the full text of instructions given to participants and screenshots, if applicable, as well as details about compensation (if any)?
    \item[] Answer: \answerNA{}
    \item[] Justification: This paper does not involve crowdsourcing or research with human subjects.

\item {\bf Institutional review board (IRB) approvals or equivalent for research with human subjects}
    \item[] Question: Does the paper describe potential risks incurred by study participants, whether such risks were disclosed to the subjects, and whether Institutional Review Board (IRB) approvals (or an equivalent approval/review based on the requirements of your country or institution) were obtained?
    \item[] Answer: \answerNA{}
    \item[] Justification: This paper does not involve crowdsourcing or research with human subjects.

\item {\bf Declaration of LLM usage}
    \item[] Question: Does the paper describe the usage of LLMs if it is an important, original, or non-standard component of the core methods in this research? Note that if the LLM is used only for writing, editing, or formatting purposes and does \emph{not} impact the core methodology, scientific rigor, or originality of the research, declaration is not required.
    \item[] Answer: \answerNA{}
    \item[] Justification: The core method development in this research does not involve LLMs as any important, original, or non-standard component.

\end{enumerate}

\fi

\end{document}